%% file: main.tex
\documentclass[11pt, a4paper,logo,  copyright, nonumbering]{paddleocr}
\usepackage[numbers]{natbib}
\usepackage{dblfloatfix}
\usepackage{ulem}
\usepackage{float}
\usepackage{tocloft}
\usepackage{caption}
\usepackage{dramatist}
\usepackage{xspace}
\usepackage{array}
\usepackage{pifont} %
\usepackage{multirow}
\usepackage{geometry}
\usepackage{makecell}
\usepackage{tcolorbox}
\usepackage{xltabular}
\usepackage{titlesec} % For \subsubsubsection support
\usepackage{longtable}
\usepackage[hang, flushmargin]{footmisc}
\usepackage{booktabs} % For better looking tables
\usepackage{xcolor}   % For coloring text
\usepackage[table]{xcolor}
\definecolor{lightpink}{RGB}{255, 182, 193}
\definecolor{iccvblue}{rgb}{0.21,0.49,0.74}

\hypersetup{
    colorlinks=true,
    linkcolor=blue,
    citecolor=blue,
    filecolor=blue,
    urlcolor=blue,
    pdfpagemode=UseNone,
    pdfstartview=FitH
}
\usepackage{threeparttable} 
\usepackage{soul}
\usepackage{amsfonts}
\usepackage{mathbbol}
\usepackage{amsmath}
\usepackage{amssymb}
\usepackage{lineno}
\usepackage{multirow}
\usepackage{adjustbox}

\usepackage{CJKutf8}
\usepackage{subcaption}
\usepackage{setspace}

\usepackage{dsfont}
\usepackage{array} %
\usepackage{tabularx} %
\usepackage{xcolor} %
\usepackage{tabularx}
\usepackage{booktabs}

\usepackage{lipsum}  %
\usepackage{multicol} %
\usepackage{makecell} 
\usepackage{listings}
\usepackage{xcolor}

\usepackage{tablefootnote}
\usepackage{hyperref}

\lstdefinelanguage{json}{
    basicstyle=\ttfamily\footnotesize,
    numbers=left,
    numberstyle=\tiny\color{gray},
    stepnumber=1,
    numbersep=8pt,
    showstringspaces=false,
    breaklines=true,
    frame=lines,
    backgroundcolor=\color{gray!10},
    morestring=[b]",
    literate=
     *{0}{{{\color{black}0}}}{1}
      {1}{{{\color{black}1}}}{1}
      {2}{{{\color{black}2}}}{1}
      {3}{{{\color{black}3}}}{1}
      {4}{{{\color{black}4}}}{1}
      {5}{{{\color{black}5}}}{1}
      {6}{{{\color{black}6}}}{1}
      {7}{{{\color{black}7}}}{1}
      {8}{{{\color{black}8}}}{1}
      {9}{{{\color{black}9}}}{1}
}

\makeatletter
\def\@BTrule[#1]{%
  \ifx\longtable\undefined
    \let\@BTswitch\@BTnormal
  \else\ifx\hline\LT@hline
    \nobreak
    \let\@BTswitch\@BLTrule
  \else
     \let\@BTswitch\@BTnormal
  \fi\fi
  \global\@thisrulewidth=#1\relax
  \ifnum\@thisruleclass=\tw@\vskip\@aboverulesep\else
  \ifnum\@lastruleclass=\z@\vskip\@aboverulesep\else
  \ifnum\@lastruleclass=\@ne\vskip\doublerulesep\fi\fi\fi
  \@BTswitch}
\makeatother

\addto\extrasenglish{
}

 {\begin{list}{}%
         {\setlength{\leftmargin}{#1}}%
         \item[]%
 }
 {\end{list}}

\reportnumber{001} %

\renewcommand{\today}{}

\title{\centering PP-OCRv6: From 1.5M to 34.5M Parameters, Surpassing Billion-Scale VLMs on OCR Tasks}

\author[*]{
\small
Yubo Zhang, Xueqing Wang, Manhui Lin, Yue Zhang, Penglongyi Deng, 
\vspace{-0.4cm}
\\
\small
Ting Sun, Tingquan Gao, Zelun Zhang, Jiaxuan Liu, Changda Zhou, 
\\
\small
Hongen Liu, Suyin Liang, Cheng Cui$^{\dagger}$, Yi Liu, Dianhai Yu, Yanjun Ma
\vspace{0.2cm}
\\
\small
\textbf{PaddlePaddle Team, Baidu Inc.}
\\
\small
$^{\dagger}$Project Leader
\\
\small
\texttt{paddleocr@baidu.com}
\vspace{0.2cm}
  \\
  {\small
  \raggedright{
  \small
  \hspace{12.6em}
  \includegraphics[height=1.0em]{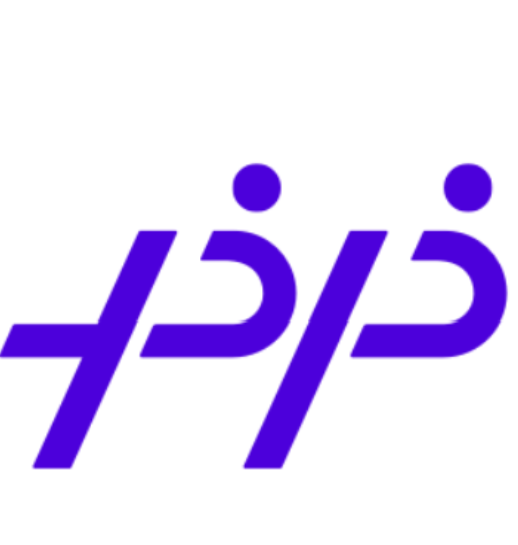} \textbf{Official Website}: \url{https://www.paddleocr.com} \\
  \hspace{-1.2em}
  \small
  \hspace{6.55em}
  \includegraphics[height=0.9em]{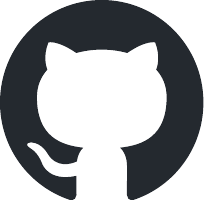} \textbf{Source Code}: \url{https://github.com/PaddlePaddle/PaddleOCR} \\
  \hspace{-1.2em}
  \small
  \hspace{6.9em}
  \includegraphics[height=1.0em]{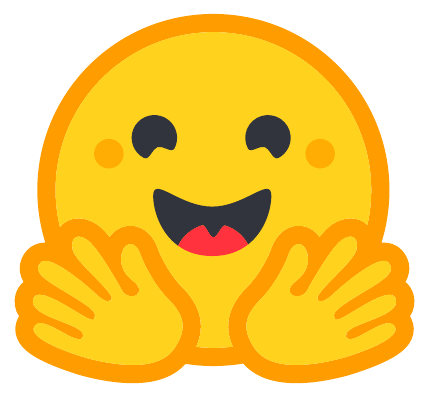} \textbf{Models}: \url{https://huggingface.co/PaddlePaddle} \\
  }

  }
}

\input{commands.tex}

\begin{abstract}

\vspace{-0.5cm} 

Vision-Language Models (VLMs) have achieved impressive results on general vision-language tasks, yet they suffer from hallucination, imprecise localization, and prohibitive computational cost when applied to dedicated OCR scenarios. This paper presents PP-OCRv6, a lightweight OCR system that combines architectural innovation with data-centric optimization. PP-OCRv6 redesigns the backbone, detection neck, and recognition neck around a unified MetaFormer-style building block with structural reparameterization, decoupling spatial token mixing from channel mixing and supporting both tasks through task-specific stride configurations. Three model tiers (medium, small, tiny) share the same block primitives, covering deployment scenarios from server to edge. On our in-house benchmarks, PP-OCRv6\_medium achieves 83.2\% recognition accuracy and 86.2\% detection Hmean, outperforming PP-OCRv5\_server by +5.1\% and +4.6\% respectively while surpassing Qwen3-VL-235B, GPT-5.5, and Gemini-3.1-Pro with orders of magnitude fewer parameters. The tiny tier achieves 3.9$\times$ faster inference than PP-OCRv5\_mobile on Intel Xeon CPU while maintaining comparable accuracy.

\end{abstract}

\begin{document}

\maketitle

% \begin{figure*}[t]
%     \centering
%     \includegraphics[width=\textwidth]{v6_images_v2/v6_images_v2_v2.png}
%     \caption{Performance comparison between PP-OCRv6, PP-OCRv5, and Vision-Language Models. Left: text recognition weighted average accuracy (\%) on our in-house benchmark. Right: text detection average Hmean (\%) on the detection benchmark. PP-OCRv6 models (blue) consistently outperform both PP-OCRv5 (gray) and VLMs (red) on both tasks, with PP-OCRv6\_medium achieving 83.2\% recognition accuracy and 86.2\% detection Hmean.}
%     \label{fig:overview_acc}
% \end{figure*}

\vspace{-0.2cm} 
\begin{figure}[h]

\makebox[0pt][l]{\hspace{-0.5cm}\includegraphics[width=1.0\textwidth]{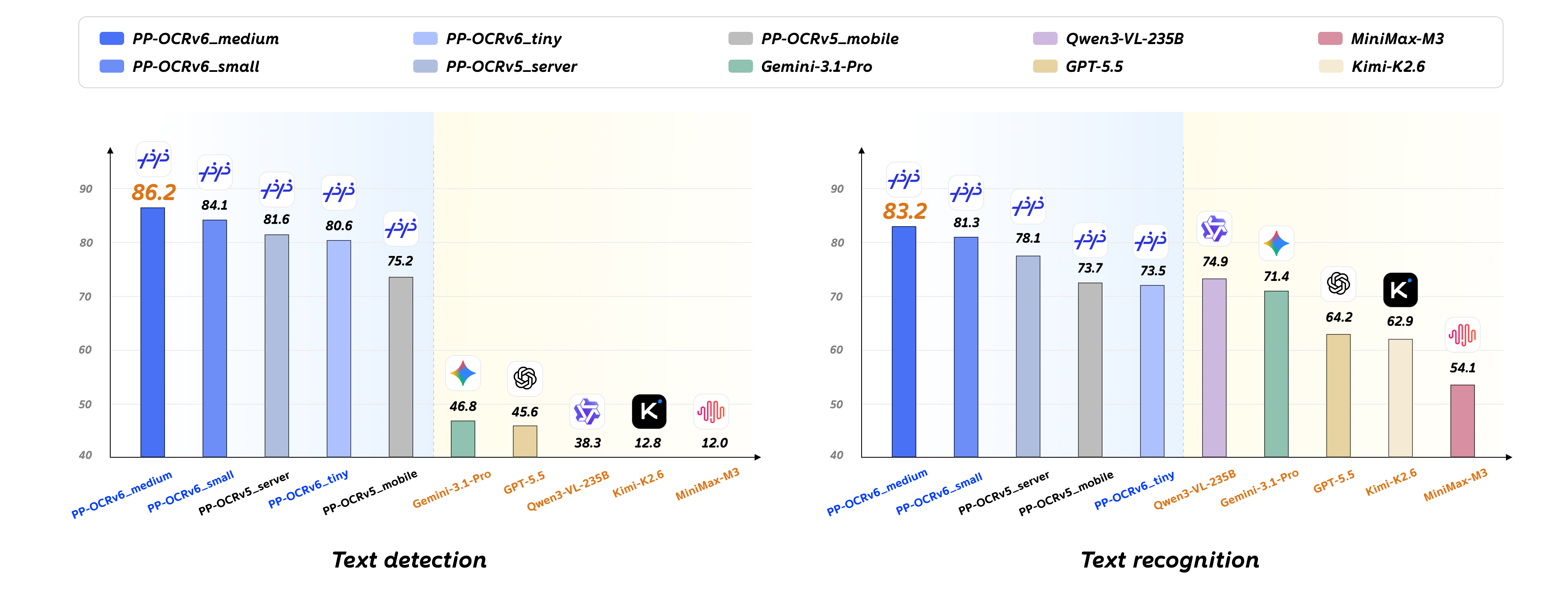}}

\caption{
    % \centering
    Performance comparison between PP-OCRv6, PP-OCRv5, and Vision-Language Models. Left: text detection average Hmean (\%) on our in-house benchmark. Right: text recognition weighted average accuracy (\%) on the recognition benchmark.  
}
\label{fig:dataset}
\end{figure}
\vspace{-0.2cm} 

% \newpage
% \setlength{\cftbeforesecskip}{6pt}   % 一级标题（section）间隔
% \setlength{\cftbeforesubsecskip}{4pt} 
% \setcounter{tocdepth}{2}
% \tableofcontents

\newpage

\section{Introduction}
\label{sec:intro}

The field of computer vision is increasingly dominated by large-scale models, particularly Vision-Language Models (VLMs), which have demonstrated impressive generalist capabilities across a wide range of tasks~\cite{achiam2023gpt,yang2025qwen3}. This trend of scaling model size has naturally extended to OCR, with modern VLM-based approaches promising an end-to-end solution for extracting text from complex, real-world images. While the performance of these large models is academically remarkable, the prevailing focus on scale often encounters significant practical limitations when deployed in real-world OCR scenarios, which demand not only high recognition accuracy but also precision, reliability, and efficiency.

A primary challenge arises from the ''generalist's dilemma.'' Unified VLM architectures, although designed for diverse tasks, often lack the specialized capabilities needed for high-precision OCR. This manifests in three critical limitations: (1) Imprecise Localization: These models often fail to produce the tight, accurate bounding boxes essential for document analysis. (2) Hallucination: In complex layouts, models can confidently generate plausible yet incorrect text, presenting a critical vulnerability for data-sensitive applications. (3) Computational Inefficiency: Massive parameter counts render these models impractical for deployment in resource-constrained environments or for services demanding high-throughput, low-latency processing.

These challenges naturally steer the focus back towards lightweight, specialized OCR systems. PaddleOCR~\cite{du2020pp} has established a strong track record in this direction, evolving through PP-OCRv2~\cite{du2021pp}, PP-OCRv3~\cite{li2022pp}, PP-OCRv4, and PP-OCRv5~\cite{cui2026pp}. The PP-OCR series has explored two complementary directions for improving accuracy. On the architecture side, PP-OCRv3 introduced SVTR-LCNet and LK-PAN, while PP-OCRv4 adopted PPHGNetV2 and LCNetV3 as server and mobile backbones respectively. On the data side, PP-OCRv5 inherited the PP-OCRv4 architecture and focused on systematic data curation along three dimensions: difficulty, accuracy, and diversity. PP-OCRv5 demonstrated that with sufficient high-quality data, the performance ceiling of a 5M-parameter model is far higher than commonly assumed.

In this work, we present PP-OCRv6, which combines both directions. While inheriting PP-OCRv5's proven data curation methodology, PP-OCRv6 introduces a fundamentally redesigned model architecture. We observe that under the same data regime, the existing backbone designs (HGNetV2 for server, LCNetV3 for mobile) have reached their structural capacity. Specifically, the MobileNet-style DW$\to$SE$\to$PW block in LCNetV3 does not exploit modern lightweight design principles such as the MetaFormer paradigm~\cite{yu2022metaformer,wang2024repvit}; the dense $9\times9$ convolutions in LKPAN can be replaced by efficient reparameterizable alternatives with equivalent receptive fields; and the dual-architecture approach (separate backbone families for server and mobile) increases engineering complexity unnecessarily.

Our main contributions are summarized as follows:
\begin{enumerate}[leftmargin=*, nosep]
    \item \textbf{Unified and Scalable Model Family:} We deliver a comprehensive, three-tier OCR model family spanning 1.5M to 34.5M parameters. The medium-tier model achieves an 86.2\% detection H-mean and 83.2\% recognition accuracy, serving as a highly efficient, production-ready infrastructure for both industrial deployment and large-scale data pipelines.

    \item \textbf{Tailored Lightweight Architectural Innovations:}  We introduce a suite of lightweight architectural components custom-designed for OCR tasks. Specifically, we propose: (i) LCNetV4, a MetaFormer-style lightweight backbone integrated with structural reparameterization; (ii) RepLKFPN(Reparameterizable Large-Kernel Feature Pyramid Network), a detection neck leveraging dilated reparameterizable depthwise convolutions for large-scale receptive fields; and (iii) EncoderWithLightSVTR, a recognition neck powered by local-global attention and additive skip connections.

    \item \textbf{Extensive Multi-Language and Scenario Generalization:}  We scale the unified model to support 50 languages and a wide range of challenging industrial scenes (e.g., digital displays, dot-matrix characters, and tire prints). This significantly enhances OCR performance in specialized scenarios that are traditionally underserved by general-purpose visual-language models.
\end{enumerate}

These results provide compelling evidence that lightweight, specialized OCR systems---when equipped with modern architecture designs and high-quality training data---offer a practical and effective alternative for production OCR deployment in the large-model era.

\begin{figure}[t]
    \centering
    \includegraphics[width=\textwidth]{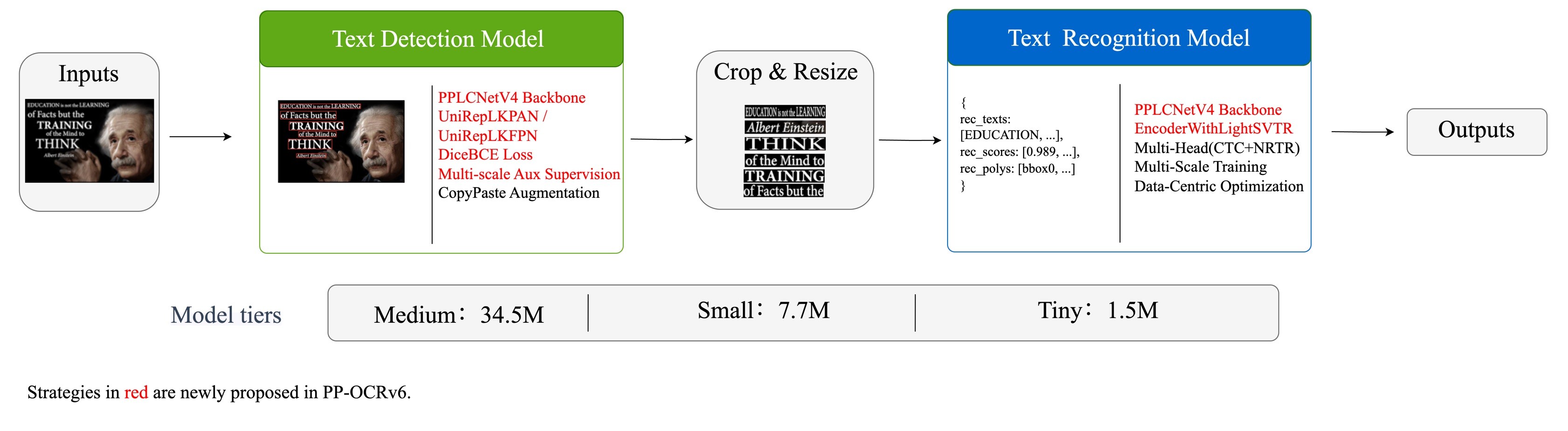}
    \caption{Overview of the PP-OCRv6 system. The pipeline consists of text detection and text recognition, both built on the unified LCNetV4 backbone. Three model variants (medium, small, tiny) share the same block primitives with different depth/width configurations.}
    \label{fig:system_overview}
\end{figure}

% \begin{figure}[t]
%     \centering
%     \includegraphics[width=\textwidth]{figuresv2/01_system_overview.pdf}
%     \caption{Overview of the PP-OCRv6 system. The pipeline consists of text detection and text recognition, both built on the unified LCNetV4 backbone. Three model variants (medium, small, tiny) share the same block primitives with different depth/width configurations.}
%     \label{fig:system_overview}
% \end{figure}

\section{Improvement Strategies}

This section presents the key technical improvements in PP-OCRv6. We first describe LCNetV4, the unified backbone (Section~\ref{sec:backbone}), then detail detection-specific (Section~\ref{sec:detection}) and recognition-specific (Section~\ref{sec:recognition}) improvements.

\subsection{LCNetV4: MetaFormer-Style Lightweight Backbone}
\label{sec:backbone}

PP-OCRv5 employs two separate backbone families: PPHGNetV2 for the server model and LCNetV3~\cite{cui2021pp} for the mobile model. This dual-architecture design increases engineering complexity in training, deployment, and maintenance. LCNetV3 follows the MobileNet-style~\cite{sandler2018mobilenetv2} convention in which depthwise convolution, squeeze-and-excitation, and pointwise convolution are stacked sequentially without explicit functional decomposition. This design conflates spatial aggregation and cross-channel interaction into a single monolithic block, limiting the flexibility to optimize each operation independently.

The MetaFormer~\cite{yu2022metaformer} framework offers an alternative perspective: an effective architecture can be decomposed into a token mixer for spatial feature aggregation and a channel mixer for per-position feature transformation. RepViT~\cite{wang2024repvit} validated this decomposition on lightweight CNNs, showing consistent accuracy gains over MobileNet-style baselines at comparable computational budgets.

\begin{figure*}[t]
    \centering
    \includegraphics[width=\textwidth]{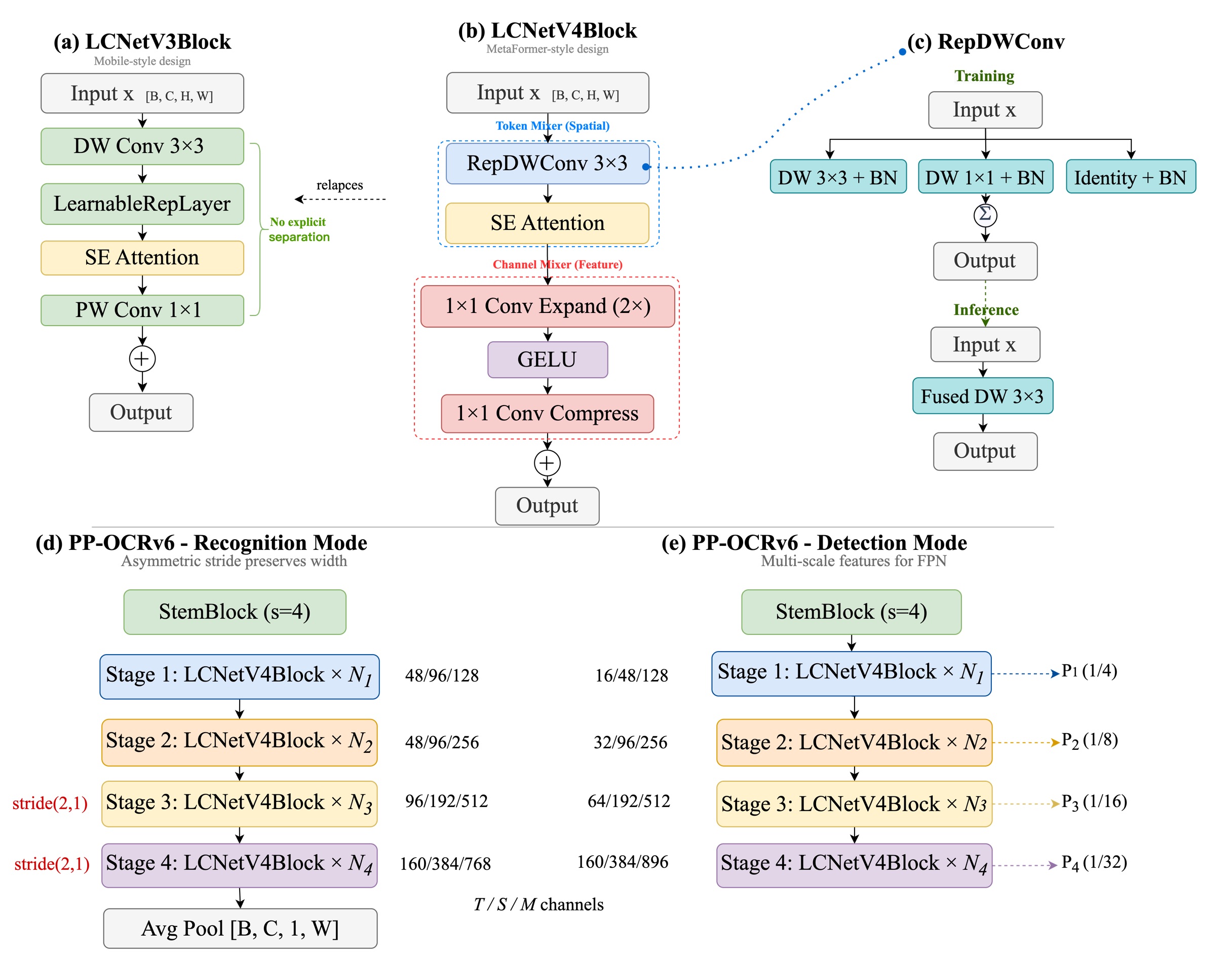}
    \caption{LCNetV4 backbone architecture. Left: block comparison between LCNetV3Block and LCNetV4Block, along with RepDWConv detail. Right: LCNetV4 in recognition mode and detection mode.}
    \label{fig:backbone}
\end{figure*}

Building on these findings, we design LCNetV4 around an explicit token mixer / channel mixer decomposition, augmented with structural reparameterization~\cite{ding2021repvgg} to enrich the training-time representational capacity without increasing inference cost. A single LCNetV4 implementation serves both text detection and text recognition through task-specific stride configurations (Figure~\ref{fig:backbone}), unifying the previously separate backbone families into one codebase.

\subsubsection{LCNetV4Block}

Following the MetaFormer paradigm, the LCNetV4Block (Figure~\ref{fig:backbone}) decomposes each layer into a token mixer followed by a channel mixer. Let $\mathbf{x} \in \mathbb{R}^{C \times H \times W}$ denote the input feature. The block computes:
\begin{align}
    \hat{\mathbf{x}} &= \text{SE}\big(\text{DW}(\mathbf{x})\big) + \mathbf{x} \label{eq:token_mixer} \\
    \mathbf{y} &= W_2\,\sigma(W_1\,\hat{\mathbf{x}}) + \hat{\mathbf{x}} \label{eq:channel_mixer}
\end{align}
where $\text{DW}(\cdot)$ is a $3\times3$ depthwise convolution (the token mixer), SE is an optional squeeze-and-excitation~\cite{hu2018squeeze} module, $W_1 \in \mathbb{R}^{2C \times C}$ and $W_2 \in \mathbb{R}^{C \times 2C}$ are $1\times1$ pointwise convolutions forming the channel mixer with expansion ratio~2, and $\sigma$ is GELU. Both stages employ identity shortcuts when the spatial resolution and channel dimension remain unchanged.

This separation has two practical advantages. First, it provides independent design knobs: the token mixer controls the spatial receptive field while the channel mixer controls the per-position representational capacity, and the two can be tuned without mutual interference. Second, it enables targeted application of structural reparameterization to the token mixer alone, as discussed below.

\paragraph{Reparameterization of the token mixer.}
The token mixer in Eq.~\eqref{eq:token_mixer} performs depthwise convolution followed by a residual addition. As shown by RepVGG~\cite{ding2021repvgg}, any convolution with an identity shortcut can be augmented with additional parallel branches during training and then merged into a single convolution at inference via algebraic equivalence. We apply this insight to the depthwise layer: during training, we maintain a $3\times3$ DW branch, a $1\times1$ DW branch, and a batch-normalized identity branch, all summed before a shared BN layer. At inference, the three branches are fused into one $3\times3$ DW convolution by zero-padding the $1\times1$ kernel and the identity to $3\times3$, then summing the weight tensors. This introduces zero additional inference cost while providing a richer optimization landscape during training through the implicit ensemble of different-scale receptive fields.

We do not apply reparameterization to the channel mixer because $1\times1$ convolutions already operate with maximal parameter efficiency per FLOP (every parameter contributes a unique cross-channel interaction) and lack spatial extent, so multi-branch augmentation offers negligible benefit. This asymmetric use of reparameterization, applying it to the spatial DW layer but not the $1\times1$ layers, follows the same design principle as RepViT~\cite{wang2024repvit}.

\paragraph{Comparison with LCNetV3.}
Table~\ref{tab:block_compare} summarizes the architectural differences. The original MobileNetV3 block in LCNetV3 places the $1\times1$ expansion before the DW convolution, coupling token mixing and channel mixing into an inseparable sequence. By contrast, LCNetV4Block moves the DW convolution to the front and applies the $1\times1$ expand-compress pair as a standalone residual channel mixer. This enables both the use of structural reparameterization on the DW layer and the BN zero-initialization on the compress layer (which makes each block behave as identity at the start of training, stabilizing convergence for deeper configurations).

\begin{table}[t]
\centering
\small
\begin{tabular}{lll}
\toprule
\textbf{Design Aspect} & \textbf{LCNetV3} & \textbf{LCNetV4 (Ours)} \\
\midrule
Architecture Paradigm & MobileNet-style & MetaFormer \\
 & (DW$\to$SE$\to$PW) & (TokenMixer + ChannelMixer) \\
Channel Interaction & Single 1$\times$1 PW Conv & Expand(2$\times$)$\to$Act$\to$Compress \\
 & & + residual connection \\
Spatial Mixing & Plain DW Conv & RepDWConv (3-branch) \\
 & & (3$\times$3 + 1$\times$1 + identity) \\
BN Initialization & Standard & Zero-init on compress BN \\
Scaling Strategy & Uniform scale factor & Per-tier depth/width config \\
\bottomrule
\end{tabular}
\caption{Block design comparison between LCNetV3 and LCNetV4.}
\label{tab:block_compare}
\end{table}

\subsubsection{Network Architecture and Scaling}

LCNetV4 is designed to serve both text detection and text recognition within a single codebase, differentiated only by stride configuration and output interface. Unlike LCNetV3, which applies a uniform scale factor to resize all channels simultaneously, LCNetV4 adopts explicit per-tier configurations with independently tuned depth and width at each stage (Table~\ref{tab:net_config}).

The two task modes differ in their spatial processing strategy. In detection mode, each stage transition applies standard stride-2 downsampling, producing multi-scale feature maps at strides 4, 8, 16, and 32 for FPN-based processing. In recognition mode, the width dimension must be preserved for sequential CTC/NRTR decoding. We address this by applying asymmetric stride $(2,1)$ at the Stage~3 and Stage~4 transitions, which halves the height at each step while leaving the width intact. After Stage~4, average pooling along the height axis produces a 1-D feature sequence of shape $[B, C, 1, W]$. This asymmetric stride mechanism is what allows a single backbone to output both 2-D spatial pyramids (for detection) and 1-D sequential features (for recognition). Table~\ref{tab:net_config} summarizes the per-tier configurations for both tasks; Figure~\ref{fig:backbone} illustrates both modes.

\begin{table}[t]
\centering
\small
\begin{tabular}{llccccc}
\toprule
\textbf{Task} & \textbf{Tier} & \textbf{Stem} & \textbf{Stage 1} & \textbf{Stage 2} & \textbf{Stage 3} & \textbf{Stage 4} \\
\midrule
\multirow{3}{*}{Det} & Tiny & Branch(16) & 2$\times$16 & 3$\times$32 & 5$\times$64 & 3$\times$160 \\
 & Small & Branch(48) & 2$\times$48 & 3$\times$96 & 5$\times$192 & 3$\times$384 \\
 & Medium & Branch(128) & 2$\times$128 & 3$\times$256 & 5$\times$512 & 3$\times$896 \\
\midrule
\multirow{3}{*}{Rec} & Tiny & Simple(48) & 1$\times$48 & 1$\times$48 & 3$\times$96 & 4$\times$160 \\
 & Small & Branch(96) & 1$\times$96 & 2$\times$96 & 7$\times$192 & 3$\times$384 \\
 & Medium & Branch(128) & 1$\times$128 & 3$\times$256 & 7$\times$512 & 3$\times$768 \\
\midrule
\multicolumn{7}{l}{\textit{Full model parameters (Det / Rec)}} \\
\multicolumn{2}{l}{\quad Tiny} & \multicolumn{5}{l}{0.43M / 1.1M \quad (end-to-end: 1.5M)} \\
\multicolumn{2}{l}{\quad Small} & \multicolumn{5}{l}{2.48M / 5.2M \quad (end-to-end: 7.7M)} \\
\multicolumn{2}{l}{\quad Medium} & \multicolumn{5}{l}{15.5M / 19M \quad (end-to-end: 34.5M)} \\
\bottomrule
\end{tabular}
\caption{LCNetV4 network configurations for detection and recognition. Depth: number of LCNetV4Blocks per stage. Width: channel dimension. Recognition mode uses asymmetric stride $(2,1)$ in Stage~3--4 to preserve width. Params denotes full model parameters (backbone + neck + head).}
\label{tab:net_config}
\end{table}

\subsection{Text Detection}
\label{sec:detection}
In PP-OCRv6, we adopt LCNetV4 as the backbone network for the text detection model. On top of the new backbone, we apply three additional strategies to further improve detection accuracy while maintaining efficiency: (1) RepLKFPN, a lightweight feature pyramid with large receptive field; (2) Auxiliary Deep Supervision for better intermediate feature learning; (3) introducing Focal Loss~\cite{lin2017focal} combined with Dice Loss~\cite{milletari2016v} for per-pixel supervision.

\begin{figure*}[t]
    \centering
    \includegraphics[width=\textwidth]{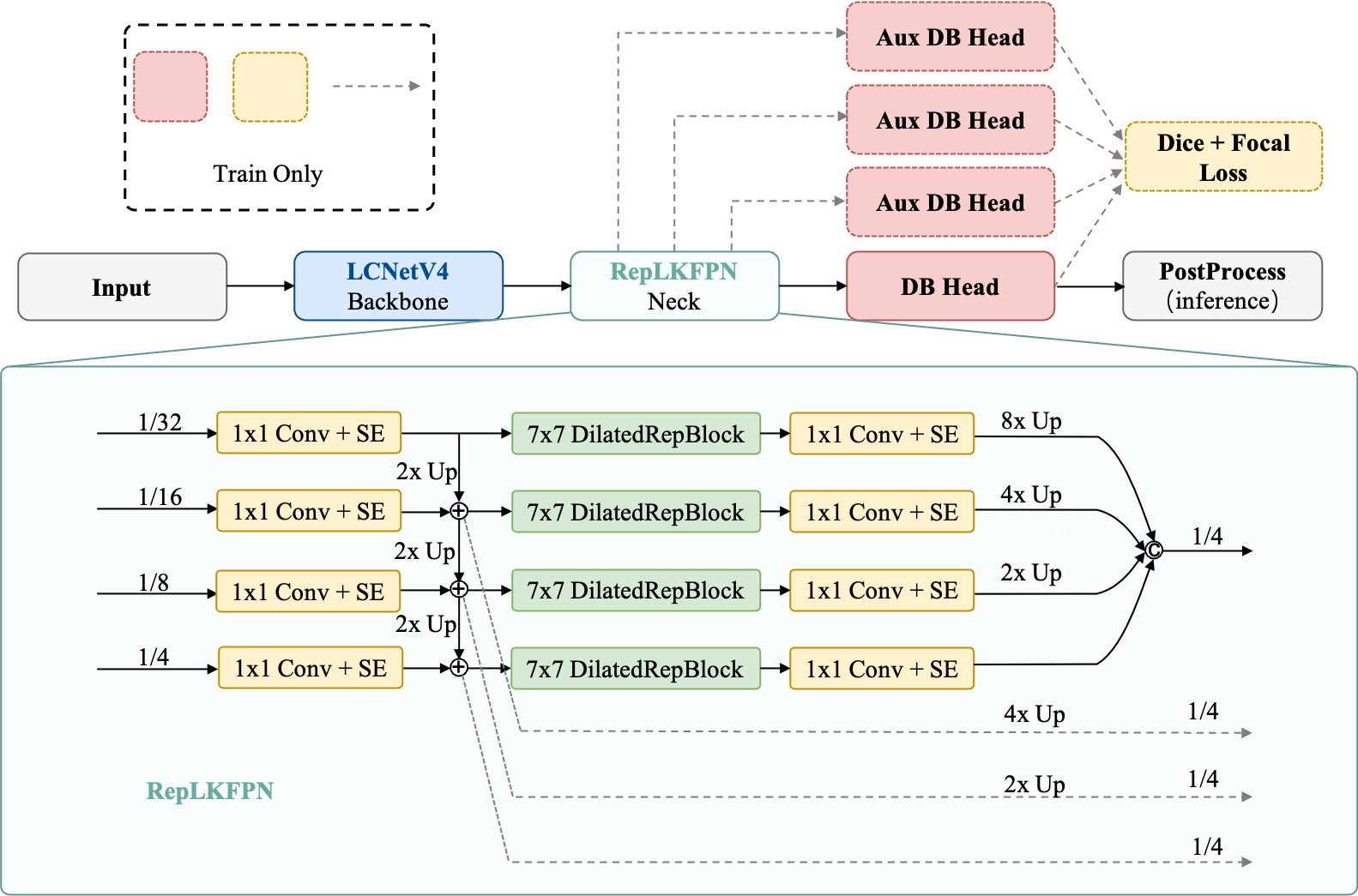}
    \caption{PP-OCRv6 text detection architecture. Top: overall pipeline with LCNetV4 backbone, RepLKFPN neck, DB head, and training-only auxiliary DB heads. Bottom: RepLKFPN performs top-down multi-scale fusion and large-kernel per-level refinement before aggregating outputs at $1/4$ resolution.}
    \label{fig:det_arch}
\end{figure*}

\subsubsection{RepLKFPN: Lightweight Large-Kernel Feature Pyramid}
The feature pyramid network (FPN) plays a critical role in fusing multi-scale features for text detection. In PP-OCRv5\_mobile\_det, RSEFPN uses standard 3$\times$3 convolutions with Squeeze-and-Excitation (SE) attention~\cite{hu2018squeeze} as the output convolution block at each pyramid level. However, these standard convolutions are parameter-heavy, while each block only provides a local 3$\times$3 receptive field.

RepLKFPN keeps the overall top-down FPN fusion structure unchanged and replaces only this per-level refinement block with a depthwise-separable design: a DilatedReparamBlock~\cite{ding2024unireplknet} followed by a 1$\times$1 pointwise convolution and SE attention. In our implementation, the DilatedReparamBlock uses a 7$\times$7 depthwise convolution as the main branch, plus three parallel dilated depthwise branches (5$\times$5 with dilation 1, 3$\times$3 with dilation 2, and 3$\times$3 with dilation 3). These branches enrich the per-level refinement block with multi-scale spatial patterns during training. At inference, all branches are merged into a single 7$\times$7 depthwise convolution via structural reparameterization~\cite{ding2021repvgg}, incurring no additional branch latency. Table~\ref{tab:fpn_params} compares the parameter count of the whole FPN neck and the local receptive field of the per-level refinement block.

\begin{table}[h]
\centering
\begin{tabular}{llll}
\toprule
\textbf{Module} & \textbf{Params} & \textbf{Per-level refinement block} & \textbf{Block RF} \\
\midrule
RSEFPN & 172K & 3$\times$3 Conv & 3$\times$3 \\
RepLKFPN (train) & 140K & DilatedReparamBlock & 7$\times$7 \\
RepLKFPN (infer) & 118K & 7$\times$7 DWConv & 7$\times$7 \\
\bottomrule
\end{tabular}
\caption{Parameter comparison of the whole FPN neck (out\_channels=96). The RF column denotes the local receptive field of the per-level refinement block.}
\label{tab:fpn_params}
\end{table}

This design reduces the whole FPN neck parameters from 172K to 118K after reparameterization, while increasing the local RF of the per-level refinement block from 3$\times$3 to 7$\times$7. This provides stronger local context modeling for fused pyramid features and benefits the detection of both large and densely-arranged text.
\subsubsection{Auxiliary Deep Supervision}
In PP-OCRv5\_mobile\_det, supervision is only applied to the final fused feature map, which means gradient signals must propagate through the entire FPN before reaching the backbone. This can lead to insufficient learning of intermediate features, particularly at deeper scales.

We introduce auxiliary prediction heads~\cite{lee2015deeply} at three FPN levels (P2, P3, P4). Each auxiliary head independently predicts the shrink map, threshold map, and binary map using the same head architecture as the main branch. During training, the auxiliary outputs are supervised with the same loss formulation as the main branch (Dice+Focal loss on the shrink map and binary map, L1 loss on the threshold map), weighted by per-scale coefficients ($\lambda_{\text{P2}}=0.4$, $\lambda_{\text{P3}}=0.3$, $\lambda_{\text{P4}}=0.2$). This deep supervision provides stronger gradient signals to intermediate features and acts as a regularizer. At inference, the auxiliary heads are removed, so the model size and speed remain unchanged.

\subsubsection{Focal Loss for Per-Pixel Supervision}
PP-OCRv5\_mobile\_det uses only DiceLoss~\cite{milletari2016v} for the probability map supervision. DiceLoss is a set-level metric that computes a single scalar measuring the global overlap between prediction and ground truth---it provides less explicit pixel-wise discrimination between easy and hard samples. Although OHEM~\cite{shrivastava2016training} is applied, it relies on per-pixel loss ranking to select hard negatives, which may limit the effectiveness of hard-sample mining because the optimization objective is dominated by global region overlap..

To address this, we introduce Focal Loss~\cite{lin2017focal} as an explicit per-pixel supervision signal and combine it with DiceLoss:
\begin{equation}
\mathcal{L}_{\text{shrink}} = \lambda_{\text{dice}} \cdot \mathcal{L}_{\text{Dice}} + \lambda_{\text{focal}} \cdot \mathcal{L}_{\text{Focal}}
\end{equation}
where $\lambda_{\text{dice}} = \lambda_{\text{focal}} = 1.0$. Focal Loss computes an independent loss value for each pixel via $-\alpha_t(1-p_t)^\gamma \log(p_t)$, providing true per-pixel supervision that DiceLoss alone cannot offer. We set $\alpha=0.25$ and $\gamma=2.5$. The two losses are complementary: DiceLoss optimizes global region overlap, while Focal Loss provides fine-grained per-pixel gradients that adaptively emphasize hard examples, enabling more precise learning at the pixel level.

\subsection{Text Recognition}
\label{sec:recognition}

The PP-OCRv6 recognition system (Figure~\ref{fig:rec_arch}) adopts a Backbone--Neck--Head pipeline. The LCNetV4 backbone encodes the input text line image ($3 \times 48 \times W$) into a compact feature map $\mathbf{F} \in \mathbb{R}^{B \times C \times 1 \times W}$, where the height has been collapsed to 1 by the asymmetric stride design described above. This feature map is then passed to the EncoderWithLightSVTR neck (for medium/small models) or directly reshaped (for tiny), before being decoded by a multi-head architecture combining CTC~\cite{hu2020gtc} and NRTR~\cite{sheng2019nrtr}. At inference, only the CTC branch is evaluated for parallel decoding; the NRTR branch participates only during training, functioning as an implicit language model that regularizes the shared encoder representations through auxiliary cross-entropy supervision with label smoothing.

\subsubsection{EncoderWithLightSVTR}

The recognition neck in PP-OCRv5 (EncoderWithSVTR) applies global self-attention to the feature sequence and then fuses the result with the input via concatenation along the channel axis, followed by a $1\times1$ projection from $2C$ back to $C$. This concatenation-based skip demands a $2C \times C$ projection matrix, which is parameter-heavy (e.g., $2 \times 120 \times 120 = 28{,}800$ parameters for the mobile model) and provides limited inductive bias about the local structure of text.

\begin{figure*}[t]
    \centering
    \includegraphics[width=\textwidth]{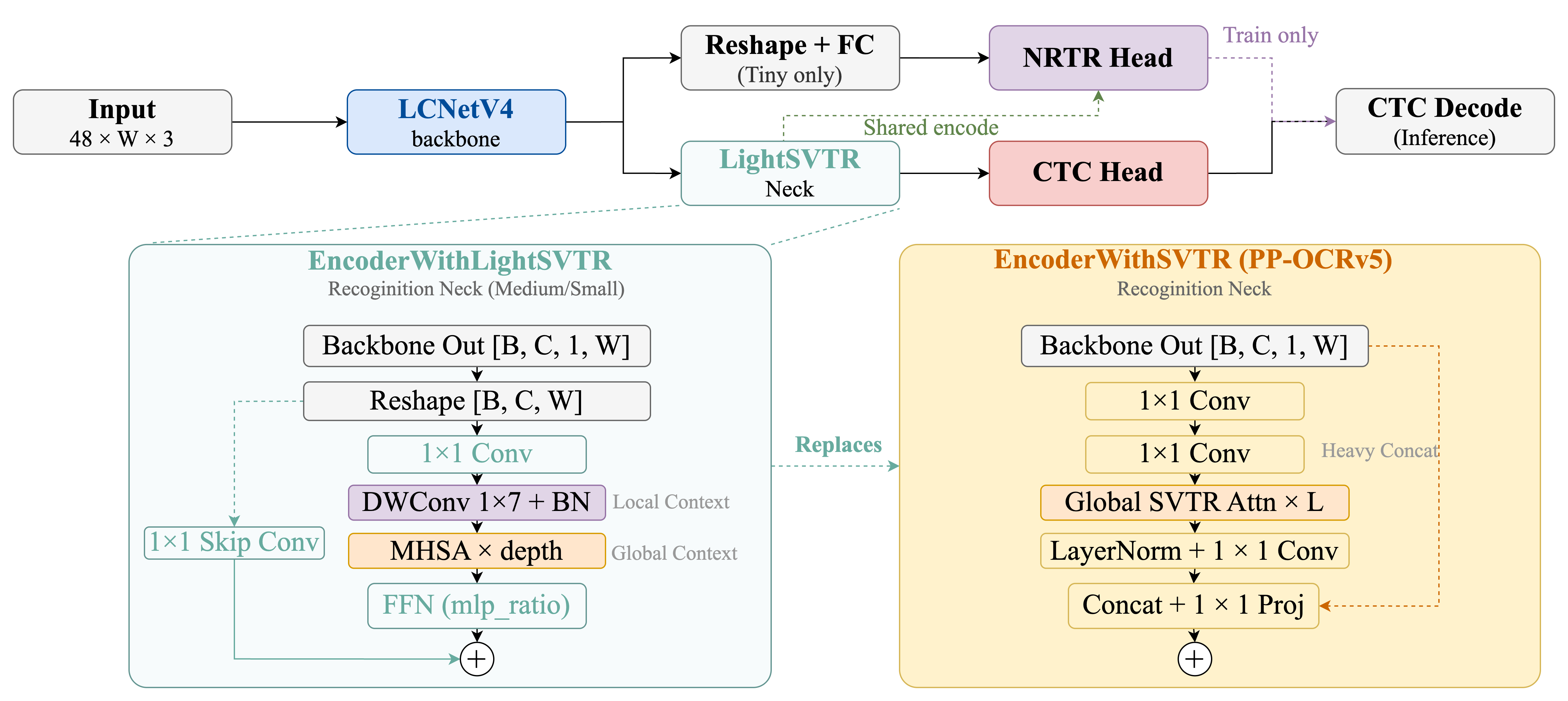}
    \caption{PP-OCRv6 text recognition architecture. Top: overall recognition pipeline where input passes through LCNetV4 backbone, then LightSVTR neck (medium/small) or Reshape+FC (tiny), feeding into CTC Head for inference and NRTR Head for training. Bottom: comparison between EncoderWithLightSVTR and EncoderWithSVTR.}
    \label{fig:rec_arch}
\end{figure*}

We propose \textbf{EncoderWithLightSVTR} (Figure~\ref{fig:rec_arch}) to address both issues simultaneously. The design rests on two observations: (1) text is inherently sequential, so local context along the horizontal axis is complementary to global attention; (2) additive skip connections are sufficient for feature fusion when dimensions match, avoiding the doubled channel width.

Concretely, the input feature map $\mathbf{F}$ is first projected by a $1\times1$ convolution to reduce the channel dimension from $C$ to a smaller $D$ (e.g., $D{=}192$ for medium). A depthwise convolution with kernel $1 \times 7$ is then applied along the width axis, injecting local positional awareness of neighboring characters before any global interaction occurs. This local-context-first design is motivated by the observation that character boundaries and inter-character spacing are highly informative for recognition, yet purely global attention treats all positions equally. The locally-primed features then pass through $L$ stacked Transformer blocks (MHSA + FFN). Finally, the output is added element-wise to a lightweight $1\times1$ skip projection of the original input:
\begin{equation}
    \mathbf{y} = \underbrace{\text{TransformerBlock}^{L}\!\big(\text{DWConv}_{1\times7}(\text{Conv}_{1\times1}(\mathbf{x}))\big)}_{\text{local-global path}} + \underbrace{\text{Conv}_{1\times1}^{\text{skip}}(\mathbf{x})}_{\text{skip path}}
    \label{eq:lightsvtr}
\end{equation}
The additive fusion costs only $C \times D$ parameters (vs. $2C^2$ for concatenation), and the DWConv adds merely $D \times 7$ depthwise parameters---negligible overhead. For the medium model we set $D{=}192$, $L{=}2$, MLP ratio 4.0; for small, $D{=}120$, $L{=}2$, ratio 2.0.

\subsubsection{Tiny Model: Minimalist Design}

The tiny recognition model (1.1M params) forgoes the encoder neck entirely. The backbone output ($\mathbb{R}^{B \times 160 \times 1 \times W}$) is directly reshaped and projected to 80 dimensions via a learned linear layer, then fed into the CTC/NRTR decoder heads. To compensate for the reduced model capacity, we employ knowledge distillation from the medium model. Since the tiny model uses a smaller character dictionary (to reduce output layer size and improve efficiency), we first train a dedicated medium teacher model with the same dictionary as the tiny student, ensuring vocabulary-consistent logit alignment. The teacher is then frozen during distillation. The training objective is:
\begin{equation}
    \mathcal{L}_{\text{tiny}} = \mathcal{L}_{\text{CTC}}^{\text{GT}} + \mathcal{L}_{\text{NRTR}}^{\text{GT}} + \lambda \cdot \mathcal{L}_{\text{KL}}^{\text{CTC}}
    \label{eq:kd_loss}
\end{equation}
where $\mathcal{L}_{\text{CTC}}^{\text{GT}}$ and $\mathcal{L}_{\text{NRTR}}^{\text{GT}}$ are standard ground-truth supervision losses for the CTC and NRTR branches respectively, and $\mathcal{L}_{\text{KL}}^{\text{CTC}}$ is a KL-divergence loss that aligns the student's CTC output probability distribution with the teacher's:
\begin{equation}
    \mathcal{L}_{\text{KL}}^{\text{CTC}} = \frac{1}{T}\sum_{t=1}^{T} D_{\text{KL}}\big(p_t^{\text{teacher}} \;\|\; p_t^{\text{student}}\big)
    \label{eq:ctc_kl}
\end{equation}
where $p_t$ denotes the per-timestep softmax distribution over the vocabulary. Because teacher and student share the identical dictionary, the CTC logits can be directly aligned without any projection or vocabulary mapping. The NRTR branch, supervised only by ground truth with label smoothing, acts as an auxiliary regularizer during training and is discarded at inference. We set $\lambda{=}1.0$ in all experiments.

% \subsubsection{Model Tier Summary}

% Table~\ref{tab:tier_summary} summarizes the architectural choices across all three tiers for both detection and recognition.

% \begin{table}[t]
% \centering
% \caption{PP-OCRv6 model tier summary. All tiers share the same LCNetV4Block primitive.}
% \label{tab:tier_summary}
% \small
% \begin{tabular}{lccc}
% \toprule
% \textbf{Component} & \textbf{Medium} & \textbf{Small} & \textbf{Tiny} \\
% \midrule
% \multicolumn{4}{l}{\textit{Detection}} \\
% \quad Backbone & LCNetV4-B & LCNetV4-S & LCNetV4-T \\
% \quad Neck & RepLKPAN (9$\times$9) & RepLKFPN (7$\times$7) & RepLKFPN (5$\times$5) \\
% \quad Head & DBHead + Aux & DBHead + Aux & DBHead + Aux \\
% \midrule
% \multicolumn{4}{l}{\textit{Recognition}} \\
% \quad Backbone & LCNetV4-B (768ch) & LCNetV4-S (384ch) & LCNetV4-T (160ch) \\
% \quad Neck & LightSVTR-192 & LightSVTR-120 & Reshape (80ch) \\
% \quad Decoder & CTC + NRTR-512 & CTC + NRTR-384 & CTC + NRTR-384 \\
% \quad Params & 19M & 5.2M & 1.1M \\
% \bottomrule
% \end{tabular}
% \end{table}

\section{Experiments}
\subsection{Experimental Setup}
\label{sec:setup}

\paragraph{Training Configuration.}
All models are implemented in PaddlePaddle and trained with Adam optimizer ($\beta_1=0.9$, $\beta_2=0.999$) using cosine learning rate schedule with linear warmup. Recognition models are trained for 100 epochs with L2 regularization ($3\times10^{-5}$). Detection models are trained for 500 epochs with L2 regularization ($1\times10^{-6}$). Multi-scale training with input sizes $[320\times32, 320\times48, 320\times64]$ is used for recognition. Detection uses $640\times640$ input with CopyPaste augmentation.

\paragraph{Data.}
PP-OCRv6 inherits the data-centric methodology from PP-OCRv5~\cite{cui2026pp}, training on the curated large-scale dataset constructed through principled difficulty filtering (confidence 0.95--0.97 sweet spot), accuracy verification, and diversity sampling (1000 K-means clusters). To support the expanded scenario coverage, we further augment the training data with both synthetic generation and real-world collection targeting industrial scenes such as digital displays, dot-matrix characters, tire parameters, and power adapter text. This shared data foundation enables fair ablation of the architecture improvements.

\subsection{Text Detection}
\subsubsection{Main Results}

Table~\ref{tab:det_main} compares detection Hmean on our in-house benchmark covering 16 categories across PP-OCRv6, PP-OCRv5, and state-of-the-art VLMs. Detailed ablation studies validating each individual strategy are provided in Appendix~\ref{sec:ablation_det}.

\begin{table*}[!htbp]
\centering
\resizebox{\textwidth}{!}{
\footnotesize
\begin{tabular}{@{}l c *{16}{c} @{}}
\toprule
\multirow{3}{*}{\textbf{Model}} & \multirow{3}{*}{\textbf{AVG}} & \multicolumn{16}{c}{\textbf{Scenario-specific Hmean (\%)}} \\
\cmidrule(l){3-18}
 & & \multicolumn{2}{c}{\textbf{Handwritten}} & \multicolumn{2}{c}{\textbf{Printed}} & \multirow{2}{*}{\textbf{TC}} & \multirow{2}{*}{\textbf{Anc.}} & \multirow{2}{*}{\textbf{JP}} & \multirow{2}{*}{\textbf{Blur}} & \multirow{2}{*}{\textbf{Emo.}} & \multirow{2}{*}{\textbf{Warp}} & \multirow{2}{*}{\textbf{Pin.}} & \multirow{2}{*}{\textbf{Art.}} & \multirow{2}{*}{\textbf{Tab.}} & \multirow{2}{*}{\textbf{Rot.}} & \multirow{2}{*}{\textbf{Indus.}} & \multirow{2}{*}{\textbf{Gen.}} \\
\cmidrule(l){3-6}
 & & \textbf{CN} & \textbf{EN} & \textbf{CN} & \textbf{EN} & & & & & & & & & & & & \\
\midrule
\multicolumn{18}{l}{\textit{Vision-Language Models}} \\
\quad Gemini-3.1-Pro & 46.8 & 53.4 & 56.5 & 47.3 & 47.6 & 39.0 & 45.8 & 38.2 & 50.0 & 68.1 & 44.6 & 40.6 & 65.2 & 26.9 & 22.1 & 52.5 & 50.2 \\
\quad GPT-5.5 & 45.6 & 42.4 & 58.5 & 50.2 & 51.9 & 35.0 & 26.7 & 42.0 & 49.1 & 97.5 & 37.7 & 36.3 & 52.0 & 71.0 & 10.0 & 36.2 & 32.6 \\
\quad Qwen3-VL-235B & 38.3 & 56.5 & 66.0 & 41.7 & 37.0 & 19.3 & 13.1 & 27.0 & 38.5 & 81.2 & 28.5 & 33.0 & \cellcolor{cyan!10}68.3 & 19.6 & 2.1 & 48.4 & 32.3 \\
\quad Kimi-K2.6 & 12.8 & 12.5 & 25.5 & 10.1 & 18.5 & 8.2 & 7.5 & 11.2 & 16.9 & 28.9 & 13.9 & 6.8 & 16.1 & 10.9 & 0.8 & 6.3 & 10.9 \\
\quad MiniMax-M3 & 12.0 & 13.7 & 19.3 & 9.8 & 14.1 & 7.7 & 11.1 & 10.6 & 16.1 & 32.8 & 12.8 & 8.5 & 16.6 & 5.5 & 0.1 & 6.4 & 6.4 \\
\midrule
\multicolumn{18}{l}{\textit{PP-OCRv5}} \\
\quad PP-OCRv5\_server & 81.6 & 80.3 & 84.1 & \cellcolor{cyan!10}94.5 & 91.7 & 81.5 & 67.6 & 77.2 & 90.1 & 96.2 & \cellcolor{cyan!10}87.6 & 67.1 & 67.3 & \cellcolor{red!10}97.1 & 80.0 & 64.3 & \cellcolor{cyan!10}79.7 \\
\quad PP-OCRv5\_mobile & 75.2 & 74.4 & 77.7 & 90.5 & 91.0 & 82.3 & 58.1 & 72.7 & 87.4 & 93.6 & 82.7 & 57.5 & 52.5 & 92.8 & 64.7 & 52.8 & 72.1 \\
\midrule
\multicolumn{18}{l}{\textit{PP-OCRv6 (Ours)}} \\
\quad \textbf{PP-OCRv6\_medium} & \cellcolor{red!10}\textbf{86.2} & \cellcolor{red!10}83.7 & 84.0 & \cellcolor{red!10}95.1 & \cellcolor{red!10}93.7 & \cellcolor{red!10}86.3 & \cellcolor{red!10}80.2 & \cellcolor{red!10}84.3 & \cellcolor{red!10}94.1 & 99.6 & \cellcolor{red!10}88.6 & \cellcolor{red!10}74.0 & \cellcolor{red!10}69.0 & \cellcolor{cyan!10}96.8 & \cellcolor{red!10}93.8 & \cellcolor{red!10}73.3 & \cellcolor{red!10}82.8 \\
\quad \textbf{PP-OCRv6\_small} & \cellcolor{cyan!10}\textbf{84.1} & \cellcolor{cyan!10}80.5 & \cellcolor{red!10}87.1 & 94.2 & \cellcolor{cyan!10}93.6 & \cellcolor{cyan!10}85.7 & \cellcolor{cyan!10}72.6 & \cellcolor{cyan!10}82.3 & \cellcolor{cyan!10}92.6 & \cellcolor{cyan!10}99.7 & \cellcolor{cyan!10}87.6 & \cellcolor{cyan!10}69.6 & 65.3 & 95.6 & \cellcolor{cyan!10}93.7 & \cellcolor{cyan!10}67.6 & 78.2 \\
\quad \textbf{PP-OCRv6\_tiny} & \textbf{80.6} & 79.4 & \cellcolor{cyan!10}85.9 & 93.1 & 92.3 & 83.7 & 63.0 & 76.6 & 89.3 & \cellcolor{red!10}99.8 & 86.1 & 59.0 & 60.1 & 94.7 & 91.0 & 62.0 & 73.8 \\
\bottomrule
\end{tabular}
}

\caption{Text detection Hmean (\%) on our in-house benchmark (16 categories). \colorbox{red!10}{Red} and \colorbox{cyan!10}{blue} cells indicate the best and second-best results per column, respectively.}
\label{tab:det_main}

\vspace{4pt}
\footnotesize
\textbf{Abbreviations}: CN: Chinese, EN: English, TC: Traditional Chinese, Anc.: Ancient Text, JP: Japanese, Emo.: Emoji, Pin.: Pinyin, Art.: Artistic Text, Tab.: Table OCR, Rot.: Rotate, Indus.: Industrial Characters, Gen.: General Scenes.
\end{table*}

PP-OCRv6 consistently improves text detection accuracy over previous PP-OCR models. PP-OCRv6\_medium achieves the best average Hmean of 86.2\%, outperforming PP-OCRv5\_server by 4.6 percentage points, with notable gains on Japanese, ancient text, rotated text, and industrial-character scenarios. PP-OCRv6\_small further reaches 84.1\% average Hmean and also surpasses PP-OCRv5\_server, while PP-OCRv6\_tiny obtains 80.6\% average Hmean, approaching the server baseline and clearly outperforming PP-OCRv5\_mobile.

Compared with VLMs, the PP-OCRv6 detection models show a substantial advantage in precise text localization. PP-OCRv6\_medium exceeds the best VLM result, Gemini-3.1-Pro, by 39.4 percentage points (86.2\% vs. 46.8\%), and even PP-OCRv6\_tiny outperforms it by 33.8 percentage points (80.6\% vs. 46.8\%). Qwen3-VL-235B, Kimi-K2.6, and MiniMax-M3 achieve average Hmean scores of only 38.3\%, 12.8\%, and 12.0\%, respectively. The consistent gap across all evaluated VLMs suggests that accurate text localization remains challenging for current general-purpose VLMs and continues to benefit from specialized OCR architectures.

\subsubsection{Robustness to Input Resolution Variation}

To evaluate detection robustness under input-resolution perturbations, we construct a scaled validation benchmark by randomly sampling 600 images from the detection validation set and resizing each image using seven scale factors ranging from 0.35× to 2.83×. The scales form a symmetric geometric progression around the original resolution, with a ratio of $\sqrt{2}$ between adjacent levels, providing approximately uniform sampling in scale space while covering both moderate and substantial resolution variations. Polygon annotations are scaled by the same factor as the image and validated before evaluation. We report Hmean at each tested scale, together with the mean Hmean, standard deviation, and coefficient of variation (CV) across all scales.

\begin{table*}[!htbp]
\centering
\resizebox{\textwidth}{!}{
\footnotesize
\begin{tabular}{lcccccccccc}
\toprule
\textbf{Model} & \textbf{0.35$\times$} & \textbf{0.50$\times$} & \textbf{0.71$\times$} & \textbf{1.00$\times$} & \textbf{1.41$\times$} & \textbf{2.00$\times$} & \textbf{2.83$\times$} & \textbf{Mean $\uparrow$} & \textbf{Std. $\downarrow$} & \textbf{CV $\downarrow$ (\%)} \\
\midrule
PP-OCRv4\_server & 55.76 & 69.75 & 71.32 & 73.30 & 73.42 & 68.39 & 57.09 & 67.00 & 6.91 & 10.31 \\
PP-OCRv4\_mobile & 48.53 & 65.51 & 68.38 & 68.56 & 69.73 & 68.56 & 64.65 & 64.84 & 6.87 & 10.60 \\
PP-OCRv5\_server & \cellcolor{cyan!10}74.70 & \cellcolor{cyan!10}82.37 & 86.30 & 85.88 & 84.17 & 79.13 & 67.28 & 79.98 & 6.41 & 8.02 \\
PP-OCRv5\_mobile & 66.94 & 73.59 & 79.20 & 81.50 & 80.77 & 75.39 & 65.81 & 74.74 & 5.91 & 7.90 \\
\midrule
\textbf{PP-OCRv6\_medium} & \cellcolor{red!10}76.29 & \cellcolor{red!10}85.00 & \cellcolor{red!10}89.04 & \cellcolor{red!10}89.72 & \cellcolor{red!10}89.69 & \cellcolor{red!10}89.04 & \cellcolor{red!10}87.94 & \cellcolor{red!10}86.67 & \cellcolor{red!10}4.50 & \cellcolor{red!10}5.19 \\
\textbf{PP-OCRv6\_small} & 71.86 & 81.12 & \cellcolor{cyan!10}86.35 & \cellcolor{cyan!10}88.52 & \cellcolor{cyan!10}88.65 & \cellcolor{cyan!10}87.75 & \cellcolor{cyan!10}86.52 & \cellcolor{cyan!10}84.40 & 5.64 & 6.68 \\
\textbf{PP-OCRv6\_tiny} & 69.21 & 78.12 & 83.64 & 84.74 & 84.92 & 84.24 & 81.81 & 80.95 & \cellcolor{cyan!10}5.27 & \cellcolor{cyan!10}6.52 \\
\bottomrule
\end{tabular}
}
\caption{Resolution robustness of text detection models. Hmean and Mean are higher-is-better, while Std. $\downarrow$ and CV $\downarrow$ are lower-is-better. \colorbox{red!10}{Red} and \colorbox{cyan!10}{blue} cells indicate the best and second-best results per column, respectively.}
\label{tab:det_resolution_robustness}
\end{table*}

Table~\ref{tab:det_resolution_robustness} shows that PP-OCRv6 is consistently more robust to resolution variation. PP-OCRv6\_medium obtains both the highest mean Hmean (86.67\%) and the lowest CV (5.19\%), while PP-OCRv6\_small and PP-OCRv6\_tiny also achieve lower CV than the PP-OCRv5 baselines. At the extreme 2.83$\times$ scale, all PP-OCRv6 models retain over 81\% Hmean, clearly outperforming PP-OCRv5\_server (67.28\%) and PP-OCRv5\_mobile (65.81\%).
\subsection{Text Recognition}
\label{sec:rec_results}

\subsubsection{Main Results}

Table~\ref{tab:rec_main} compares recognition accuracy on our in-house benchmark (15 categories) across PP-OCRv6, PP-OCRv5, and state-of-the-art VLMs.

\begin{table*}[!htbp]
\centering
\resizebox{\textwidth}{!}{
\footnotesize
\begin{tabular}{@{}l c *{15}{c} @{}}
\toprule
\multirow{3}{*}{\textbf{Model}} & \multirow{3}{*}{\textbf{W-Avg}} & \multicolumn{15}{c}{\textbf{Scenario-specific Accuracy (\%)}} \\
\cmidrule(l){3-17}
 & & \multicolumn{2}{c}{\textbf{Handwritten}} & \multicolumn{2}{c}{\textbf{Printed}} & \multirow{2}{*}{\textbf{TC}} & \multirow{2}{*}{\textbf{Anc.}} & \multirow{2}{*}{\textbf{JP}} & \multirow{2}{*}{\textbf{Conf.}} & \multirow{2}{*}{\textbf{Spec.}} & \multirow{2}{*}{\textbf{Gen.}} & \multirow{2}{*}{\textbf{Pin.}} & \multirow{2}{*}{\textbf{Art.}} & \multirow{2}{*}{\textbf{Indus.}} & \multirow{2}{*}{\textbf{Screen}} & \multirow{2}{*}{\textbf{Card}} \\
\cmidrule(l){3-6}
 & & \textbf{CN} & \textbf{EN} & \textbf{CN} & \textbf{EN} & & & & & & & & & & & \\
\midrule
\multicolumn{17}{l}{\textit{Vision-Language Models}} \\
\quad GPT-5.5 & 64.2 & 19.2 & 56.9 & 75.7 & 82.2 & 57.5 & 63.7 & 58.6 & 49.1 & 48.3 & 67.7 & 50.4 & 53.0 & 62.4 & 67.7 & 71.1 \\
\quad Qwen3-VL-235B & 74.9 & 49.7 & \cellcolor{red!10}73.2 & 82.3 & 86.2 & 76.4 & 33.6 & 66.2 & 56.1 & 49.0 & 82.5 & \cellcolor{cyan!10}76.5 & \cellcolor{cyan!10}69.6 & 74.7 & 73.8 & 78.7 \\
\quad Kimi-K2.6 & 62.9 & 31.0 & 58.4 & 76.8 & 80.9 & 62.7 & 16.5 & 54.1 & 43.5 & 38.0 & 68.0 & 45.2 & 59.9 & 57.1 & 58.4 & 68.4 \\
\quad MiniMax-M3 & 54.1 & 15.5 & 60.3 & 63.5 & 81.5 & 53.2 & 2.2 & 43.7 & 42.2 & 42.8 & 53.8 & 50.3 & 44.3 & 44.1 & 56.6 & 67.0 \\
\quad Gemini-3.1-Pro & 71.4 & 46.4 & \cellcolor{cyan!10}73.0 & 80.0 & \cellcolor{cyan!10}90.5 & 69.5 & 18.0 & 67.2 & 54.4 & 50.3 & 74.6 & 75.9 & 63.1 & 69.1 & 73.2 & 75.9 \\
\midrule
\multicolumn{17}{l}{\textit{PP-OCRv5 Series}} \\
\quad PP-OCRv5\_server & 78.1 & 58.0 & 59.6 & 90.1 & 85.1 & 74.7 & 60.4 & 73.7 & 59.4 & 56.8 & \cellcolor{cyan!10}86.5 & 74.4 & 64.0 & 70.2 & 68.1 & \cellcolor{cyan!10}87.6 \\
\quad PP-OCRv5\_mobile & 73.7 & 41.7 & 50.9 & 86.0 & 86.0 & 72.0 & 57.8 & 75.8 & 55.7 & 54.8 & 80.7 & 72.5 & 54.0 & 59.3 & 57.6 & 81.7 \\
\midrule
\multicolumn{17}{l}{\textit{PP-OCRv6 (Ours)}} \\
\quad \textbf{PP-OCRv6\_medium} & \cellcolor{red!10}\textbf{83.2} & \cellcolor{red!10}62.1 & 67.8 & \cellcolor{red!10}91.5 & \cellcolor{red!10}94.1 & \cellcolor{red!10}78.6 & \cellcolor{red!10}72.4 & \cellcolor{red!10}90.5 & \cellcolor{red!10}64.9 & \cellcolor{red!10}61.7 & \cellcolor{red!10}87.5 & \cellcolor{red!10}78.1 & \cellcolor{red!10}71.2 & \cellcolor{red!10}77.4 & \cellcolor{red!10}82.5 & \cellcolor{red!10}88.1 \\
\quad \textbf{PP-OCRv6\_small} & \cellcolor{cyan!10}\textbf{81.3} & \cellcolor{cyan!10}57.6 & 61.1 & \cellcolor{cyan!10}90.5 & 93.3 & \cellcolor{cyan!10}77.0 & \cellcolor{cyan!10}71.1 & 88.2 & \cellcolor{cyan!10}64.1 & \cellcolor{cyan!10}60.2 & 85.7 & 75.9 & 68.4 & \cellcolor{cyan!10}76.4 & \cellcolor{cyan!10}79.7 & 86.9 \\
\quad \textbf{PP-OCRv6\_tiny} & \textbf{73.5} & 40.1 & 39.3 & 86.7 & 88.4 & 65.0 & 68.4 & \cellcolor{cyan!10}89.8 & 52.3 & 57.1 & 78.0 & 65.4 & 54.7 & 62.1 & 71.2 & 80.5 \\
\bottomrule
\end{tabular}
}
\caption{Text recognition accuracy (\%) on our in-house benchmark (15 categories). \colorbox{red!10}{Red} and \colorbox{cyan!10}{blue} cells indicate the best and second-best results per column, respectively.}
\label{tab:rec_main}

\vspace{4pt}
\footnotesize
\textbf{Abbreviations}: CN: Chinese, EN: English, TC: Traditional Chinese, Anc.: Ancient Text, JP: Japanese, Conf.: Confusable Characters, Spec.: Special Characters, Gen.: General Scenes, Pin.: Pinyin, Art.: Artistic Text, Indus.: Industrial Characters, Screen: Screen Displays, Card: Physical Card Surfaces.
\end{table*}

PP-OCRv6\_medium achieves 83.2\% weighted average, outperforming PP-OCRv5\_server by +5.1\% absolute with fewer parameters. The gains are particularly pronounced on categories requiring strong multilingual and contextual understanding: Japanese (+16.8\%), Ancient text (+12.0\%), and Screen displays (+14.4\%). PP-OCRv6\_small (5.2M) surpasses PP-OCRv5\_server (21M) by +3.2\% with 4$\times$ fewer parameters, demonstrating the architectural efficiency of LCNetV4.

Compared to VLMs, the performance gap is substantial. The best-performing VLM (Qwen3-VL-235B, 74.9\%) trails PP-OCRv6\_medium by 8.3\%, despite having approximately 6800$\times$ more parameters. Even PP-OCRv6\_tiny (1.1M) exceeds four of the five VLMs in weighted average, trailing only Qwen3-VL-235B by 1.4\%. VLMs exhibit particularly poor performance on low-resource categories such as Ancient text (Gemini: 18.0\%, MiniMax: 2.2\%) and Handwritten Chinese (GPT-5.5: 19.2\%), indicating that these categories require specialized supervision.

\subsubsection{Hallucination Evaluation}

A critical advantage of specialized OCR models over VLMs is the absence of text hallucination---generating text not present in the input image. Table~\ref{tab:hallucination} evaluates this on our curated hallucination benchmark, where accuracy measures the rate of correct outputs without hallucinated content.

\begin{table}[!htbp]
\centering
\small
\begin{tabular}{llc}
\toprule
\textbf{Type} & \textbf{Model} & \textbf{Accuracy (\%)} \\
\midrule
\multirow{4}{*}{VLMs} & Qwen3-VL-235B & 80.56 \\
 & GPT-5.5 & 78.00 \\
 & Kimi-K2.6 & 85.00 \\
 & MiniMax-M3 & 72.60 \\
\midrule
\multirow{3}{*}{PP-OCRv6} & \textbf{PP-OCRv6\_medium} & \textbf{93.20} \\
 & \textbf{PP-OCRv6\_small} & 88.20 \\
 & \textbf{PP-OCRv6\_tiny} & 86.80 \\
\bottomrule
\end{tabular}
\caption{Hallucination evaluation. Higher accuracy indicates less hallucination.}
\label{tab:hallucination}
\end{table}

PP-OCRv6\_medium achieves 93.2\% accuracy on the hallucination benchmark, compared to 85.0\% for Kimi-K2.6, 80.6\% for Qwen3-VL-235B, and 72.6\% for MiniMax-M3. This gap reflects a structural difference: PP-OCRv6's CTC+NRTR decoding architecture produces outputs grounded in visual features, whereas VLMs' auto-regressive generation may produce plausible but non-existent text. This characteristic is particularly important for data-sensitive applications such as financial documents, medical records, and legal text extraction.

\subsubsection{Robustness to Crop Margin Variation}

In practical OCR pipelines, the text detection module produces bounding boxes with varying margins around the actual text content. Depending on the detection model's behavior and post-processing parameters, the cropped text image may be tightly clipped to the character boundaries or include substantial background padding. A robust recognition model should produce consistent predictions regardless of how much margin surrounds the text. To evaluate this, we measure recognition consistency across different crop margins: for each test sample, we generate multiple crops with progressively looser boundaries and report the percentage of samples that yield identical predictions across all margin settings (Table~\ref{tab:crop_robust}).

\begin{table}[!htbp]
\centering
\small
\begin{tabular}{lcc}
\toprule
\textbf{Model} & \textbf{Consistency (\%)} & \textbf{$\Delta$ vs. v5} \\
\midrule
PP-OCRv3\_mobile & 31.05 & -- \\
PP-OCRv4\_server & 45.69 & -- \\
PP-OCRv4\_mobile & 43.41 & -- \\
PP-OCRv5\_server & 54.82 & -- \\
PP-OCRv5\_mobile & 57.74 & -- \\
\midrule
\textbf{PP-OCRv6\_medium} & \textbf{75.32} & \textbf{+20.5} \\
\textbf{PP-OCRv6\_small} & 67.80 & +10.1 \\
\textbf{PP-OCRv6\_tiny} & 44.80 & -- \\
\bottomrule
\end{tabular}
\caption{Crop margin robustness (\%). Higher is better. Measures the percentage of samples with consistent recognition results across varying crop boundaries. PP-OCRv6 demonstrates substantially improved robustness to bounding box variations.}
\label{tab:crop_robust}
\end{table}

PP-OCRv6\_medium achieves 75.32\% consistency, a +20.5\% improvement over PP-OCRv5\_server (54.82\%). This improvement suggests that LCNetV4's MetaFormer architecture and LightSVTR's local-global attention mechanism produce feature representations that are less sensitive to input crop variations. The trend is consistent across the series: v3 (31.05\%) $\to$ v4 (45.69\%) $\to$ v5 (54.82\%) $\to$ v6 (75.32\%), showing steady progress with each generation.

\subsection{End-to-End System Inference Speed}
\label{sec:speed}

We benchmark the end-to-end OCR pipeline (detection + recognition) on 200 images comprising both general scene and document images. The measurement includes image I/O from disk, pre-processing, model inference, and post-processing. The software environment is PaddlePaddle 3.2.1, ONNX Runtime 1.23.2 (CUDA 12.6), OpenVINO 2025.0, and TensorRT 8.6.1 (CUDA 11.8, paddle2onnx 2.0.2rc3). Results are summarized in Table~\ref{tab:speed}.

\begin{table*}[h]
\centering
\small
\resizebox{\textwidth}{!}{
\begin{tabular}{llccccccc}
\toprule
\textbf{Hardware} & \textbf{Backend} & \textbf{PP-OCRv6\_medium} & \textbf{PP-OCRv6\_small} & \textbf{PP-OCRv6\_tiny} & \textbf{PP-OCRv5\_server} & \textbf{PP-OCRv5\_mobile} & \textbf{PP-OCRv4\_mobile} \\
\midrule
\multirow{2}{*}{NVIDIA A100} & PaddlePaddle & 0.29 & 0.25 & 0.13 & 0.32 & 0.25 & 0.14 \\
 & TensorRT & -- & 0.32 & 0.16 & -- & 0.33 & 0.16 \\
\midrule
\multirow{3}{*}{NVIDIA V100} & PaddlePaddle & 0.72 & 0.49 & 0.21 & 0.66 & 0.50 & 0.25 \\
 & ONNX Runtime & 0.67 & 0.53 & 0.29 & 0.77 & 0.46 & 0.27 \\
 & TensorRT & 0.77 & 0.60 & 0.23 & 0.73 & 0.59 & 0.27 \\
\midrule
\multirow{3}{*}{Intel Xeon 8350C} & PaddlePaddle & 2.05 & 0.79 & 0.32 & 2.04 & 0.80 & 0.62 \\
 & OpenVINO & 1.40 & 0.59 & 0.20 & 7.30 & 0.78 & 0.60 \\
 & ONNX Runtime & 3.31 & 0.61 & 0.22 & 6.36 & 0.61 & 0.49 \\
\midrule
\multirow{2}{*}{Apple M4} & PaddlePaddle & 8.82 & 3.07 & 0.96 & $>$10 & 5.82 & 5.65 \\
 & ONNX Runtime & 5.55 & 1.29 & 0.35 & 7.20 & 1.10 & 1.02 \\
\bottomrule
\end{tabular}
}
\caption{End-to-end OCR pipeline inference speed (s/image).}
\label{tab:speed}
\end{table*}

Table~\ref{tab:speed} reveals several trends. First, PP-OCRv6\_medium consistently matches or outperforms PP-OCRv5\_server across all tested configurations: 1.1$\times$ faster on A100 (0.29s vs. 0.32s), 1.15$\times$ on V100 ONNX Runtime (0.67s vs. 0.77s), and 5.2$\times$ on Intel Xeon OpenVINO (1.40s vs. 7.30s). Second, PP-OCRv6\_small matches PP-OCRv5\_mobile in latency on most platforms while delivering higher accuracy; it is 1.9$\times$ faster on Apple M4 PaddlePaddle (3.07s vs. 5.82s) and 1.3$\times$ on Intel Xeon OpenVINO (0.59s vs. 0.78s). Third, PP-OCRv6\_tiny is the fastest model across all platforms, achieving 6.1$\times$ speedup over PP-OCRv5\_mobile on Apple M4 PaddlePaddle (0.96s vs. 5.82s), 3.9$\times$ on Intel Xeon OpenVINO (0.20s vs. 0.78s), and reaching 0.13s on A100. Overall, the depthwise-separable architecture of LCNetV4 translates into favorable latency characteristics particularly on CPU and under optimized inference backends such as OpenVINO.

\section{Conclusion}
\label{sec:conclusion}

We present PP-OCRv6, a lightweight OCR system that advances accuracy through the combination of architectural innovation and data-centric optimization. Experiments demonstrate that PP-OCRv6\_medium achieves 83.2\% recognition accuracy (+5.1\% over PP-OCRv5\_server) and 86.2\% detection Hmean (+4.6\% over PP-OCRv5\_server), while PP-OCRv6\_tiny (0.43M detection, 1.1M recognition) approaches PP-OCRv5\_server in detection and remains competitive for recognition with an order of magnitude fewer parameters. Compared to VLMs, PP-OCRv6 achieves higher accuracy with 6800$\times$ fewer parameters and produces substantially lower hallucination rates (93.2\% vs. 80.6\% for Qwen3-VL-235B). These results confirm that lightweight specialized OCR systems, when equipped with modern architecture designs and curated training data, remain the practical choice for production deployment in the large-model era.

In future work, we plan to release a PP-OCRv6\_large model with greater capacity, higher accuracy, and broader language coverage beyond the current 50 languages.

{
    \small
    \bibliography{main}
}

\newpage
\appendix

\section*{Appendix}
\label{sec:appendix}

\section{Language Support}
\label{sec:language_support}

Table~\ref{tab:lang_support} compares language coverage across PP-OCR versions. PP-OCRv3/v4 support only Simplified Chinese and English; PP-OCRv5 adds Traditional Chinese and Japanese; PP-OCRv6 medium/small further incorporate 46 Latin-script languages by adding approximately 200 diacritical characters to the dictionary. The tiny model omits Japanese to avoid the $\sim$4,000 Kanji/Kana entries that would disproportionately enlarge its 1.1M-parameter output layer.

\begin{table}[h]
\centering
\small
\resizebox{0.9\textwidth}{!}{
\begin{tabular}{lcp{8.5cm}}
\toprule
\textbf{Model} & \textbf{\# Lang} & \textbf{Supported Languages} \\
\midrule
PP-OCRv3\_mobile\_rec & 2 & Simplified Chinese, English \\
\midrule
PP-OCRv4\_mobile\_rec & 2 & Simplified Chinese, English \\
\midrule
PP-OCRv4\_server\_rec & 2 & Simplified Chinese, English \\
\midrule
PP-OCRv5\_server\_rec & 4 & Simplified Chinese, Traditional Chinese, English, Japanese \\
\midrule
PP-OCRv5\_mobile\_rec & 4 & Simplified Chinese, Traditional Chinese, English, Japanese \\
\midrule
PP-OCRv6\_medium\_rec & 50 & Simplified Chinese, Traditional Chinese, English, Japanese, French, German, Italian, Spanish, Portuguese, Dutch, Polish, Romanian, Czech, Swedish, Norwegian, Danish, Finnish, Hungarian, Turkish, Vietnamese, Indonesian, Malay, Azerbaijani, Afrikaans, Bosnian, Croatian, Welsh, Estonian, Irish, Icelandic, Kurdish, Lithuanian, Latvian, Maltese, Maori, Occitan, Slovak, Slovenian, Albanian, Swahili, Tagalog, Uzbek, Latin, Serbian (Latin), Catalan, Basque, Galician, Luxembourgish, Romansh, Quechua \\
\midrule
PP-OCRv6\_small\_rec & 50 & Simplified Chinese, Traditional Chinese, English, Japanese, French, German, Italian, Spanish, Portuguese, Dutch, Polish, Romanian, Czech, Swedish, Norwegian, Danish, Finnish, Hungarian, Turkish, Vietnamese, Indonesian, Malay, Azerbaijani, Afrikaans, Bosnian, Croatian, Welsh, Estonian, Irish, Icelandic, Kurdish, Lithuanian, Latvian, Maltese, Maori, Occitan, Slovak, Slovenian, Albanian, Swahili, Tagalog, Uzbek, Latin, Serbian (Latin), Catalan, Basque, Galician, Luxembourgish, Romansh, Quechua \\
\midrule
PP-OCRv6\_tiny\_rec & 49 & Simplified Chinese, Traditional Chinese, English, French, German, Italian, Spanish, Portuguese, Dutch, Polish, Romanian, Czech, Swedish, Norwegian, Danish, Finnish, Hungarian, Turkish, Vietnamese, Indonesian, Malay, Azerbaijani, Afrikaans, Bosnian, Croatian, Welsh, Estonian, Irish, Icelandic, Kurdish, Lithuanian, Latvian, Maltese, Maori, Occitan, Slovak, Slovenian, Albanian, Swahili, Tagalog, Uzbek, Latin, Serbian (Latin), Catalan, Basque, Galician, Luxembourgish, Romansh, Quechua \\
\bottomrule
\end{tabular}
}
\caption{Language support comparison across PP-OCR recognition models. ``\# Lang'' indicates the total number of languages supported by a single unified model.}
\label{tab:lang_support}
\end{table}

To validate the multilingual capability, we evaluate PP-OCRv6 on dedicated English and Latin-script benchmarks (Table~\ref{tab:latin_eval}). PP-OCRv6\_medium achieves 88.4\% on the English benchmark and 88.0\% on the Latin benchmark, outperforming both PP-OCRv5's language-specific models (en\_PP-OCRv5\_mobile: 86.0\%, latin\_PP-OCRv5\_mobile: 81.4\%) and the PP-OCRv5 main models (server: 79.5\%, mobile: 78.3\% on English) while using a single unified model rather than separate per-language models.

\begin{table}[h]
\centering
\small
\begin{tabular}{lcc}
\toprule
\textbf{Model} & \textbf{English (\%)} & \textbf{Latin (\%)} \\
\midrule
PP-OCRv5\_server & 79.5 & -- \\
PP-OCRv5\_mobile & 78.3 & -- \\
en\_PP-OCRv5\_mobile & 86.0 & -- \\
latin\_PP-OCRv5\_mobile & 77.8 & 81.4 \\
\midrule
\textbf{PP-OCRv6\_medium} & \textbf{88.4} & \textbf{88.0} \\
\textbf{PP-OCRv6\_small} & 86.3 & 84.0 \\
\textbf{PP-OCRv6\_tiny} & 77.3 & 63.8 \\
\bottomrule
\end{tabular}
\caption{Recognition accuracy (\%) on English and Latin-script evaluation benchmarks. PP-OCRv6 uses a single unified model for all languages, while PP-OCRv5 requires separate language-specific models for optimal English/Latin performance.}
\label{tab:latin_eval}
\end{table}

\newpage
\section{Speed Comparison of Text Detection Models}
\label{sec:det_speed}

We further evaluate the inference speed of PP-OCRv6 and PP-OCRv5 text detection models using the PaddleX Inference Benchmark platform. The benchmark is conducted on a server equipped with Intel Xeon Gold 6271C CPUs and NVIDIA Tesla V100-SXM2-32GB GPUs. GPU latency is measured with the Paddle inference backend, while CPU ONNX latency is measured with the PaddleX ONNX Runtime backend.

For each exported detection model, the platform runs batch-size-one inference on square inputs from 512 to 2048 pixels. All latency numbers are averaged over 30 measured iterations after 15 warm-up iterations. Since the benchmark records latency only, this section analyzes inference speed rather than accuracy.

\begin{figure*}[h]
    \centering
    \begin{subfigure}[t]{0.48\textwidth}
        \centering
        \includegraphics[width=\textwidth]{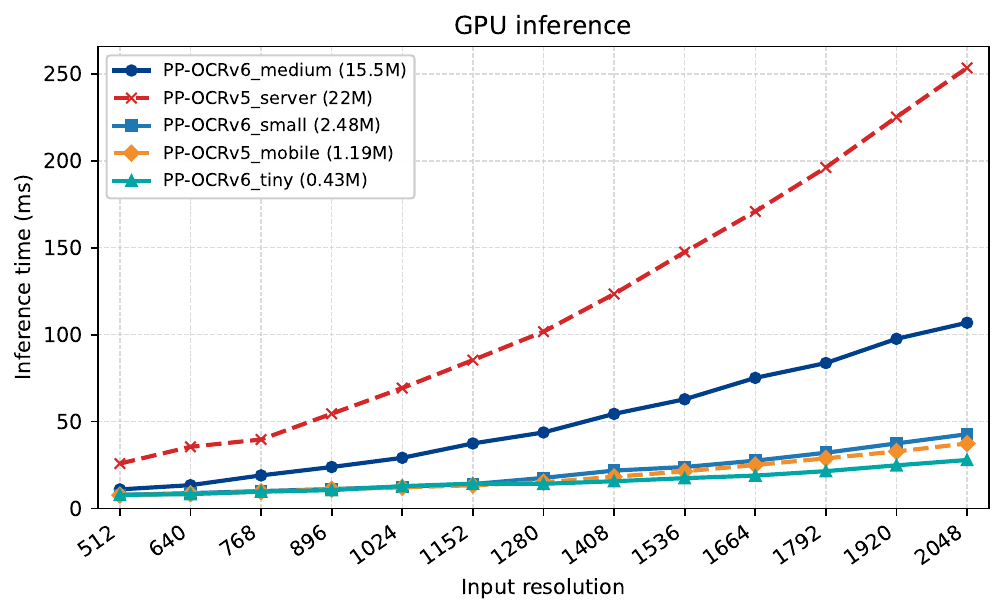}
        \caption{GPU inference.}
        \label{fig:det_speed_gpu}
    \end{subfigure}
    \hfill
    \begin{subfigure}[t]{0.48\textwidth}
        \centering
        \includegraphics[width=\textwidth]{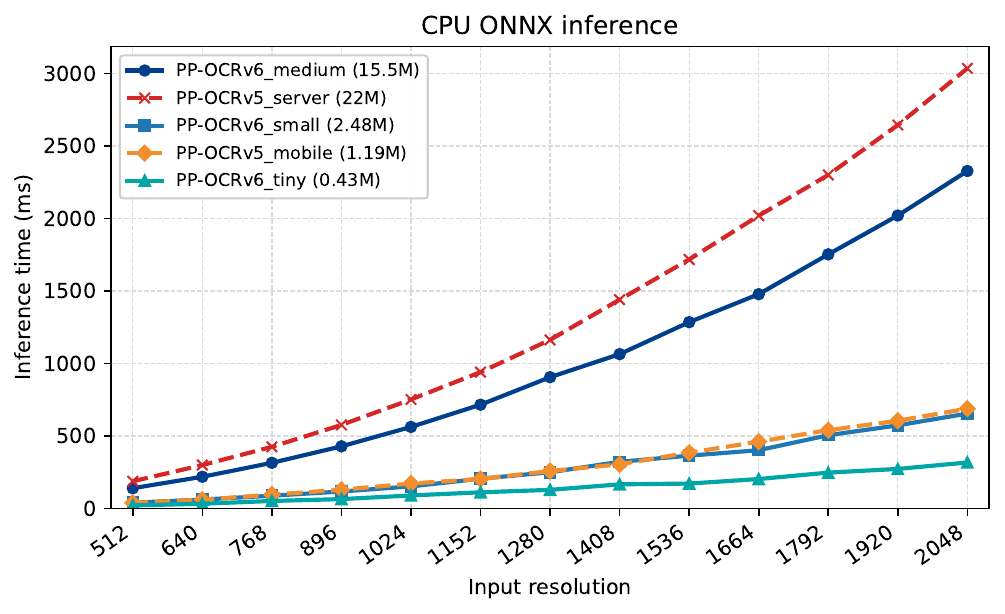}
        \caption{CPU ONNX inference.}
        \label{fig:det_speed_cpu_onnx}
    \end{subfigure}
    \caption{Inference time comparison of PP-OCRv6 and PP-OCRv5 text detection models under different input resolutions.}
    \label{fig:det_speed_comparison}
\end{figure*}

As shown in Figure~\ref{fig:det_speed_comparison}, PP-OCRv6\_medium is consistently faster than PP-OCRv5\_server under both backends. At the 2048 input resolution, PP-OCRv6\_medium takes 106.89 ms on GPU compared with 253.52 ms for PP-OCRv5\_server, corresponding to a 2.37$\times$ speedup. Under CPU ONNX inference, PP-OCRv6\_medium takes 2327.23 ms compared with 3034.93 ms for PP-OCRv5\_server, yielding a 1.30$\times$ speedup.

The lighter PP-OCRv6 tiers provide different speed-capacity trade-offs. PP-OCRv6\_small is close to PP-OCRv5\_mobile in GPU latency at 2048 resolution (42.70 ms vs. 37.36 ms) and is slightly faster under CPU ONNX inference (654.25 ms vs. 687.98 ms). PP-OCRv6\_tiny achieves the lowest CPU ONNX latency among all compared models, reaching 317.06 ms at 2048 resolution. On GPU, however, the tiny model is not always faster than the small model; for example, at 1024 resolution PP-OCRv6\_small takes 12.48 ms while PP-OCRv6\_tiny takes 12.83 ms. This indicates that practical GPU latency is affected not only by model size, but also by operator fusion, parallelism, kernel scheduling, and hardware utilization.

Across both backends, increasing the input resolution from 512 to 2048 leads to a monotonic increase in latency for all models, and the gap between model tiers becomes more pronounced at high resolution. The PP-OCRv5\_server model shows the steepest growth, while the PP-OCRv6 family maintains a more favorable speed-capacity balance across resolutions.

\newpage
\section{Ablation Studies}
\label{sec:ablation_all}

\subsection{LCNetV4 Backbone}
\label{sec:ablation_backbone}

We isolate the contribution of each backbone design choice by progressively upgrading from PP-OCRv5\_mobile (LCNetV3, scale=0.95) to the final LCNetV4. All models use the same EncoderWithSVTR neck and are trained on a representative subset (approximately 20\% of the full training data) for 50 epochs to enable rapid iteration. Recognition accuracy is measured on the in-house benchmark subset.

\begin{table}[h]
\centering
\small
\begin{tabular}{lcc}
\toprule
\textbf{Configuration} & \textbf{Acc. (\%)} & \textbf{$\Delta$} \\
\midrule
LCNetV3 (PP-OCRv5\_mobile baseline) & 76.01 & -- \\
+ MetaFormer-style block (Token/Channel Mixer separation) & 78.24 & +2.23 \\
+ RepDWConv in token mixer & 78.30 & +0.06 \\
+ Multi-branch StemBlock & 78.96 & +0.66 \\
+ Per-tier depth/width configuration & 79.21 & +0.25 \\
+ BN zero-init on channel mixer compress & 79.35 & +0.14 \\
\bottomrule
\end{tabular}
\caption{Backbone ablation on the recognition task. Each row adds one design change to the previous configuration. The neck (EncoderWithSVTR) and decoder are held constant throughout.}
\label{tab:ablation_backbone}
\end{table}

The MetaFormer-style decomposition contributes the largest single gain (+2.23\%), confirming that separating spatial aggregation from channel interaction yields substantially better representations under the same parameter budget. The multi-branch StemBlock adds +0.66\% by providing richer low-level features through parallel MaxPool and convolution paths. RepDWConv contributes a modest +0.06\% in accuracy; its primary value is zero-cost inference regularization through the implicit multi-scale ensemble during training. The explicit per-tier depth/width design (+0.25\%) and BN zero-init (+0.14\%) provide cumulative gains that reflect improved training dynamics for the deeper LCNetV4 configurations.

\subsection{Text Detection}
\label{sec:ablation_det}

To verify the effectiveness of each proposed strategy, we first conduct ablation experiments on one-fifth of the full detection training data for rapid iteration, starting from the PP-OCRv5\_mobile\_det baseline. After validating the architecture and loss components under this reduced-data setting, we train the final configuration with the full training data. Results are shown in Table~\ref{tab:det_ablation}.

\begin{table}[h]
\centering
\small
\begin{tabular}{lccc}
\toprule
\textbf{Configuration} & \textbf{Params} & \textbf{Hmean (\%)} & \textbf{$\Delta$} \\
\midrule
\multicolumn{4}{l}{\textit{Ablation with 1/5 training data}} \\
Baseline (PP-OCRv5\_mobile\_det) & 1.19M & 77.84 & -- \\
+ LCNetV4 Backbone & 2.53M & 79.37 & +1.53 \\
+ RepLKFPN & 2.48M & 79.75 & +0.38 \\
+ Auxiliary Deep Supervision & -- & 80.28 & +0.53 \\
+ Focal Loss & -- & 81.43 & +1.15 \\
\midrule
+ Full Data Training & -- & 84.12 & +2.69 \\
\bottomrule
\end{tabular}
\caption{Ablation study for PP-OCRv6\_small text detection. Each row adds one strategy on top of the previous configuration.}
\label{tab:det_ablation}
\end{table}

As shown in Table~\ref{tab:det_ablation}, upgrading the backbone from LCNetV3 to LCNetV4 brings a significant improvement of 1.53\% Hmean, indicating improved feature extraction capability of the new backbone. Replacing RSEFPN with RepLKFPN further improves Hmean by 0.38\% while slightly reducing model parameters from 2.53M to 2.48M, demonstrating the efficiency of the depthwise-separable large-kernel design. Auxiliary deep supervision and Focal Loss contribute complementary gains of 0.53\% and 1.15\%, respectively. With the full training data, the final model reaches 84.12\% Hmean, improving over the reduced-data configuration by 2.69\% and over the PP-OCRv5\_mobile\_det baseline by 6.28\%.

\subsection{Text Recognition}
\label{sec:ablation}

We evaluate the recognition-specific design choices (neck and decoder) on the final LCNetV4 backbone. All experiments use the same training subset and schedule as the backbone ablation above.

\begin{table}[h]
\centering
\small
\begin{tabular}{lcc}
\toprule
\textbf{Configuration} & \textbf{Acc. (\%)} & \textbf{$\Delta$} \\
\midrule
\multicolumn{3}{l}{\textit{Neck design}} \\
EncoderWithSVTR (PP-OCRv5, concat skip) & 79.35 & -- \\
EncoderWithLightSVTR (additive skip only) & 79.77 & +0.42 \\
EncoderWithLightSVTR (+ DWConv $1{\times}7$) & 80.24 & +0.89 \\
No neck (reshape + FC, tiny reference) & 74.52 & $-$4.83 \\
\midrule
\multicolumn{3}{l}{\textit{Decoder configuration (with LightSVTR neck)}} \\
CTC only & 79.08 & -- \\
CTC + NRTR-384 (multi-head) & 80.24 & +1.16 \\
CTC + NRTR-512 (medium only) & 80.61 & +1.53 \\
\midrule
\multicolumn{3}{l}{\textit{Knowledge distillation (tiny model, no neck)}} \\
Tiny without distillation & 71.83 & -- \\
Tiny + CTC KL distillation ($\lambda{=}1.0$) & 74.52 & +2.69 \\
\bottomrule
\end{tabular}
\caption{Recognition neck and decoder ablation. The backbone (LCNetV4-small) is fixed. Each section varies one component while holding others constant.}
\label{tab:ablation_rec}
\end{table}

Replacing the concatenation-based skip in EncoderWithSVTR with a lightweight additive connection improves accuracy by +0.42\% while reducing neck parameters. Adding the DWConv $1{\times}7$ local context layer before global attention provides a further +0.47\%, confirming that explicit local horizontal modeling is complementary to global self-attention for sequential text features. Removing the neck entirely causes a 4.83\% drop, indicating that sequence modeling in the neck is essential for medium/small capacity models.

For the decoder, the multi-head CTC+NRTR design outperforms CTC-only by +1.16\%, as the NRTR branch acts as an implicit language model that regularizes the shared encoder during training without adding inference cost. Increasing the NRTR dimension from 384 to 512 for the medium model yields an additional +0.37\%.

For the tiny model, CTC logit distillation from the vocabulary-matched medium teacher provides +2.69\% accuracy, effectively recovering much of the capacity gap introduced by removing the neck.

\newpage
\section{Qualitative Analysis of Text Detection}
\label{sec:det_qualitative}

\begin{figure*}[h]
    \centering
    \includegraphics[width=\textwidth]{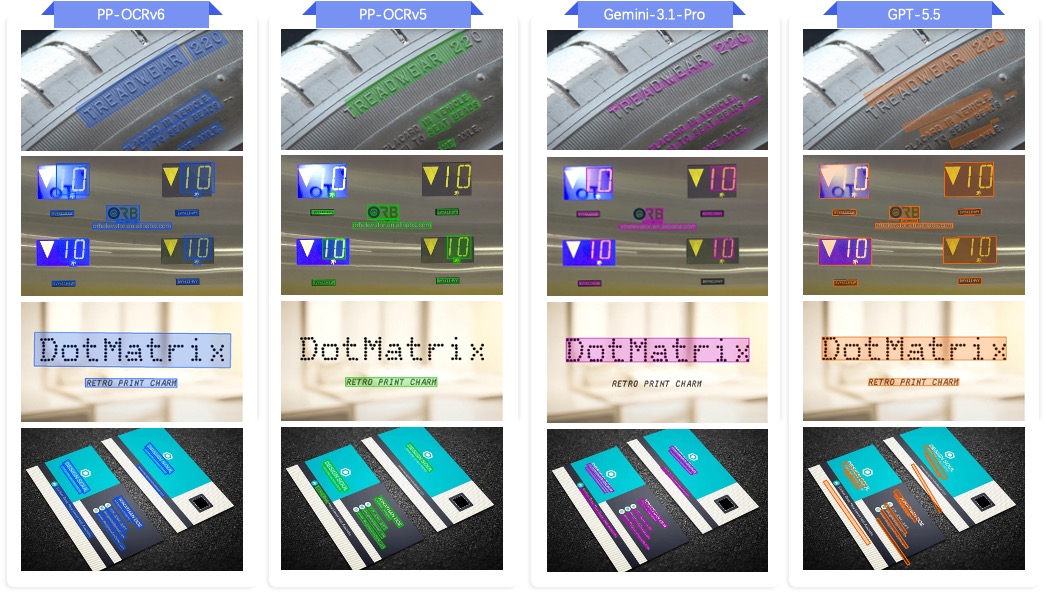}
    \caption{Qualitative comparison of text detection results. Columns from left to right: PP-OCRv6\_medium, PP-OCRv5\_server, Gemini-3.1-Pro, and GPT-5.5. Colored polygons indicate detected text regions.}
    \label{fig:det_vis}
\end{figure*}

Figure~\ref{fig:det_vis} presents qualitative detection results across four representative challenging scenarios. We compare PP-OCRv6\_medium with PP-OCRv5\_server and two state-of-the-art VLMs (Gemini-3.1-Pro and GPT-5.5).

The four rows cover curved text on a tire surface, dense small text on an electronic display, artistic/decorative fonts, and business cards, respectively. PP-OCRv6\_medium consistently produces complete and accurate polygon detections across all scenarios. For the curved tire text, PP-OCRv5 misses several text instances, while both VLMs produce overly tight bounding boxes that fail to cover the full text extent. In the electronic display scene, PP-OCRv5 misses some text elements, Gemini-3.1-Pro fails to detect small-sized characters, and GPT-5.5 generates false positives by enclosing non-text regions such as icons and symbols. For the artistic font scene, both PP-OCRv5 and Gemini-3.1-Pro miss text instances. In the business card scene, PP-OCRv5 exhibits missed detections, Gemini-3.1-Pro produces misaligned bounding boxes that do not correspond to the actual text positions, and GPT-5.5 uses axis-aligned boxes that poorly fit the oriented text lines.

These visualizations illustrate the different failure modes of each approach: PP-OCRv5 primarily suffers from missed detections in challenging scenarios, while VLMs exhibit inaccurate localization---either producing overly tight boxes, misaligned regions, or false positives on non-text content. These observations are consistent with the quantitative gaps in Table~\ref{tab:det_main}.

\newpage
\section{Hallucination Visualization Comparison with VLMs}
\label{sec:hallucination_vis}

\begin{figure*}[h]
    \centering
    \includegraphics[width=0.9\textwidth]{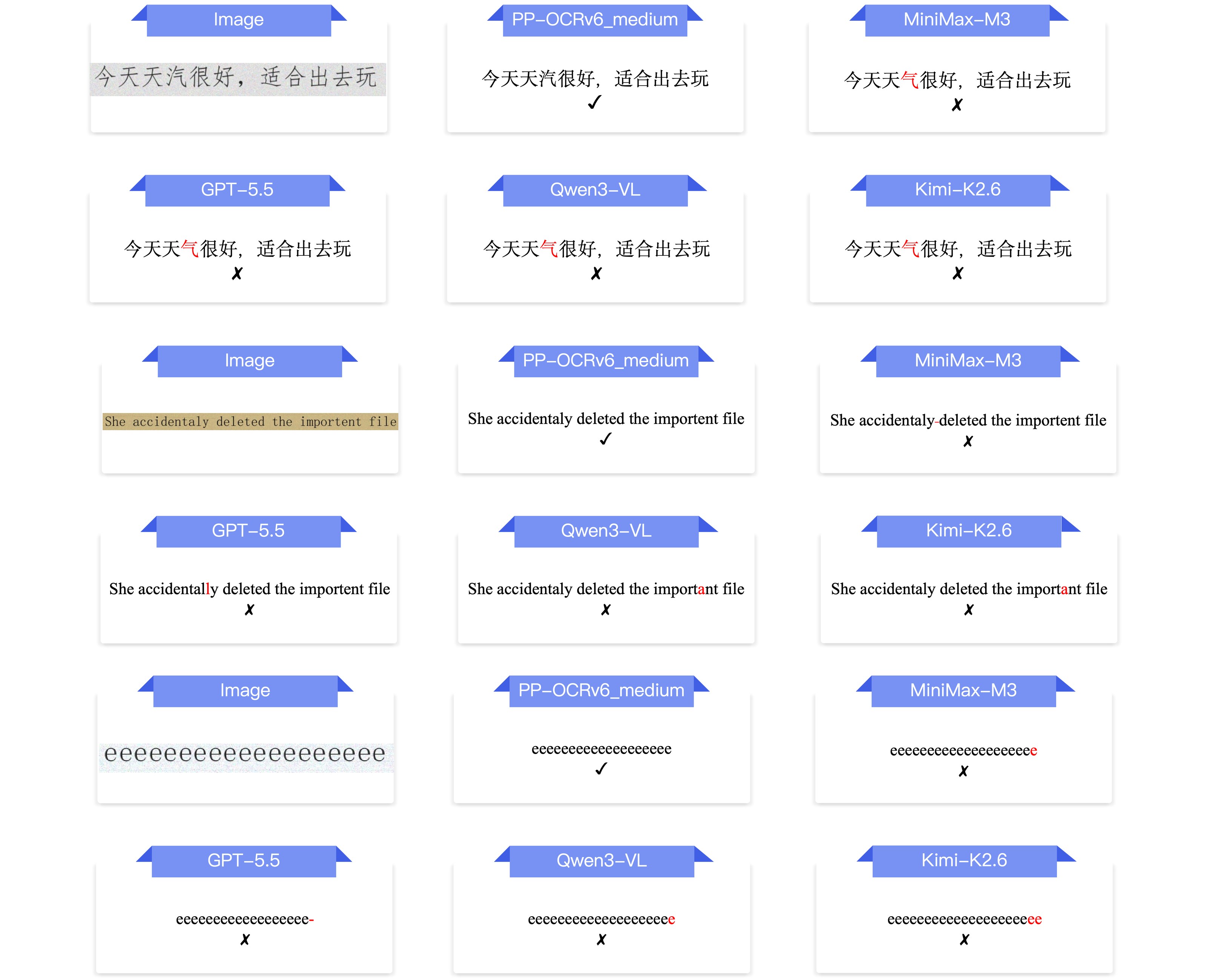}
    \caption{Hallucination comparison between PP-OCRv6\_medium and VLMs. For each example, the input image (left) contains text with non-standard spelling or repeated characters. PP-OCRv6\_medium faithfully reproduces the exact visual content (\checkmark), while VLMs (MiniMax-M3, GPT-5.5, Qwen3-VL-235B, Kimi-K2.6) modify the text based on linguistic priors (\ding{55}), introducing hallucinated corrections not present in the image. Red characters highlight discrepancies from the ground truth.}
    \label{fig:hallucination_vis}
\end{figure*}

Figure~\ref{fig:hallucination_vis} presents qualitative examples comparing PP-OCRv6\_medium with several VLMs on our hallucination evaluation set. The key observation is that VLMs tend to ``correct'' what they perceive as spelling or grammatical errors in the source image, producing text that is linguistically plausible but factually inconsistent with the visual input. In contrast, PP-OCRv6\_medium faithfully reproduces the exact text content---including deliberate misspellings, non-standard character usage, and repeated characters---without injecting linguistic priors.

This behavior reflects a fundamental architectural difference: PP-OCRv6's CTC+NRTR decoding is grounded in frame-level visual features without auto-regressive language modeling, producing outputs strictly conditioned on the input image. VLMs, by contrast, rely on strong language priors that can override visual evidence when the two conflict.

\newpage
\section{Qualitative Comparison with PP-OCRv5}
\label{sec:qualitative}

\begin{figure*}[h]
\centering
\includegraphics[width=1.0\textwidth]{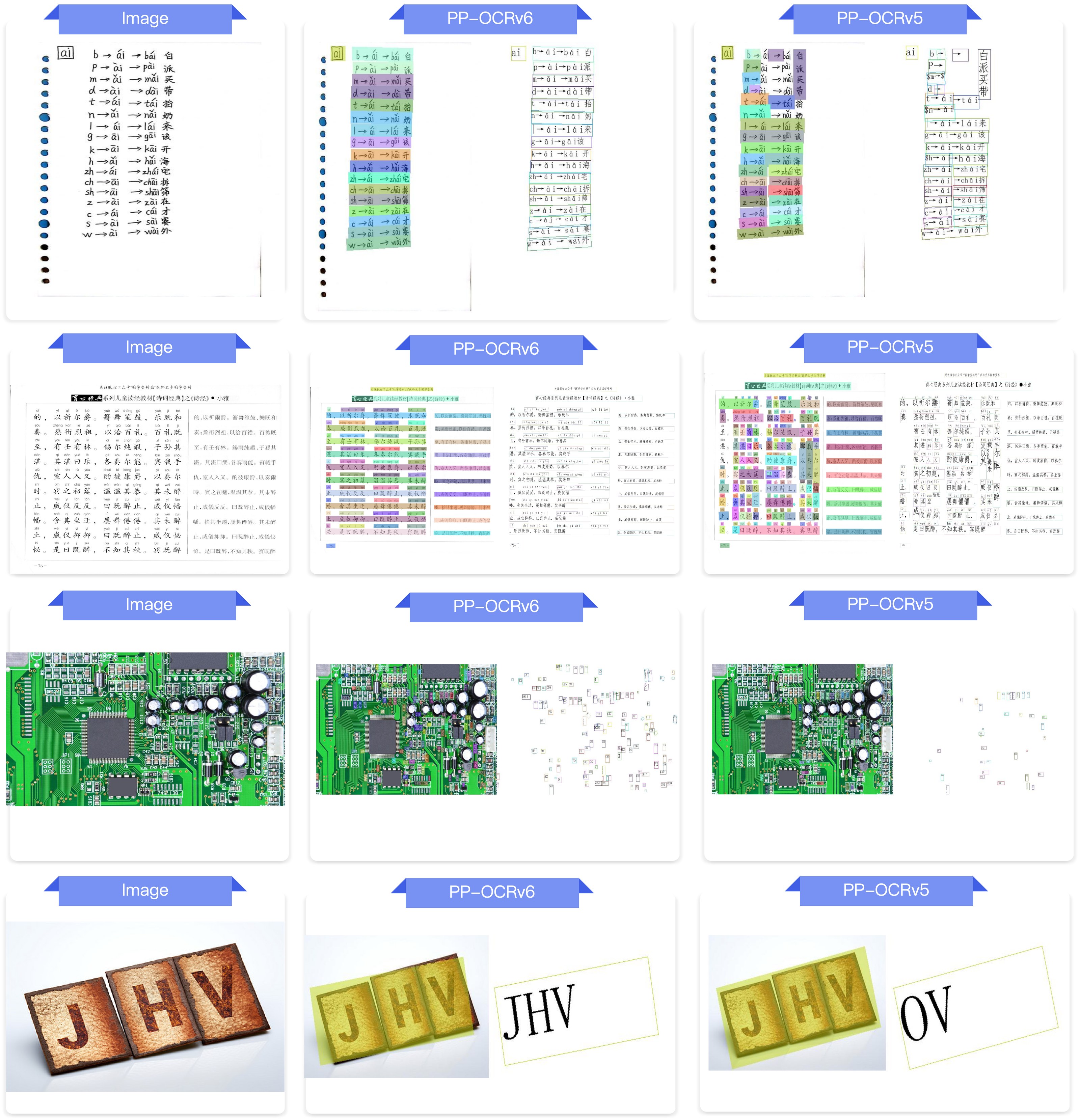}
\caption{Qualitative comparison between PP-OCRv6 and PP-OCRv5 (a).}
\label{fig:case1}
\end{figure*}

\begin{figure*}[h]
\centering
\includegraphics[width=1.0\textwidth]{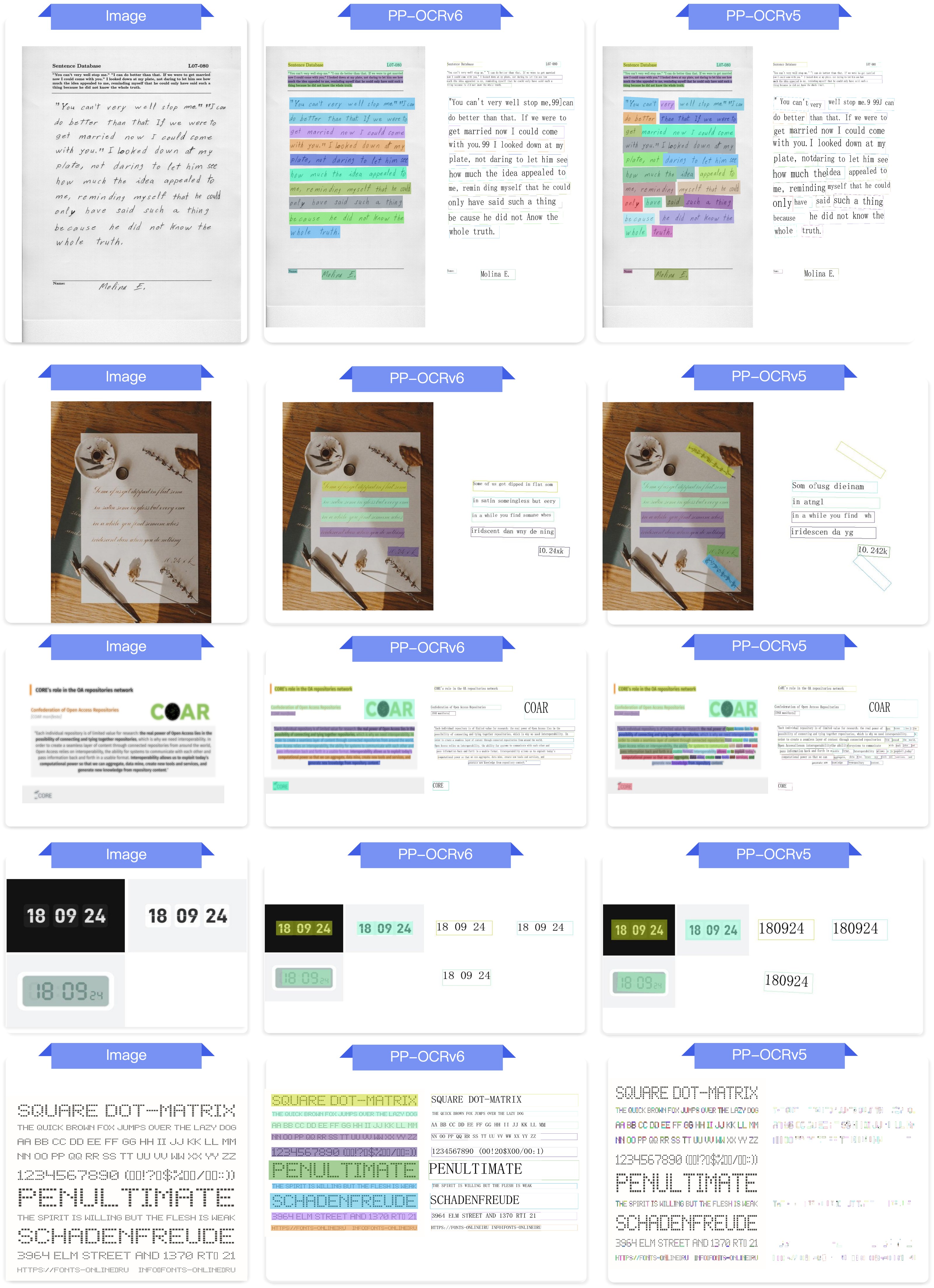}
\caption{Qualitative comparison between PP-OCRv6 and PP-OCRv5 (b).}
\label{fig:case2}
\end{figure*}

\begin{figure*}[h]
\centering
\includegraphics[width=1.0\textwidth]{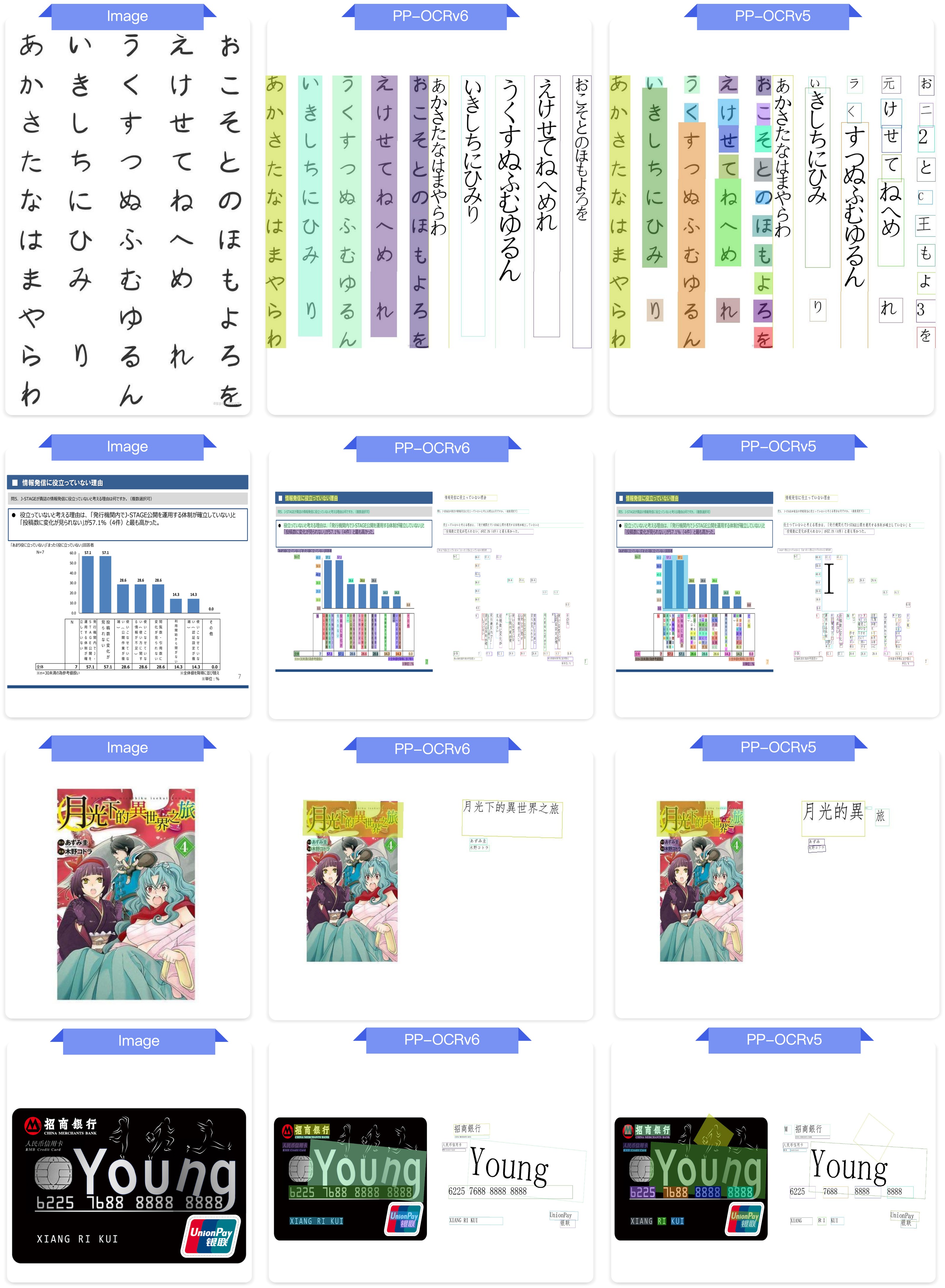}
\caption{Qualitative comparison between PP-OCRv6 and PP-OCRv5 (c).}
\label{fig:case3}
\end{figure*}

\begin{figure*}[h]
\centering
\includegraphics[width=1.0\textwidth]{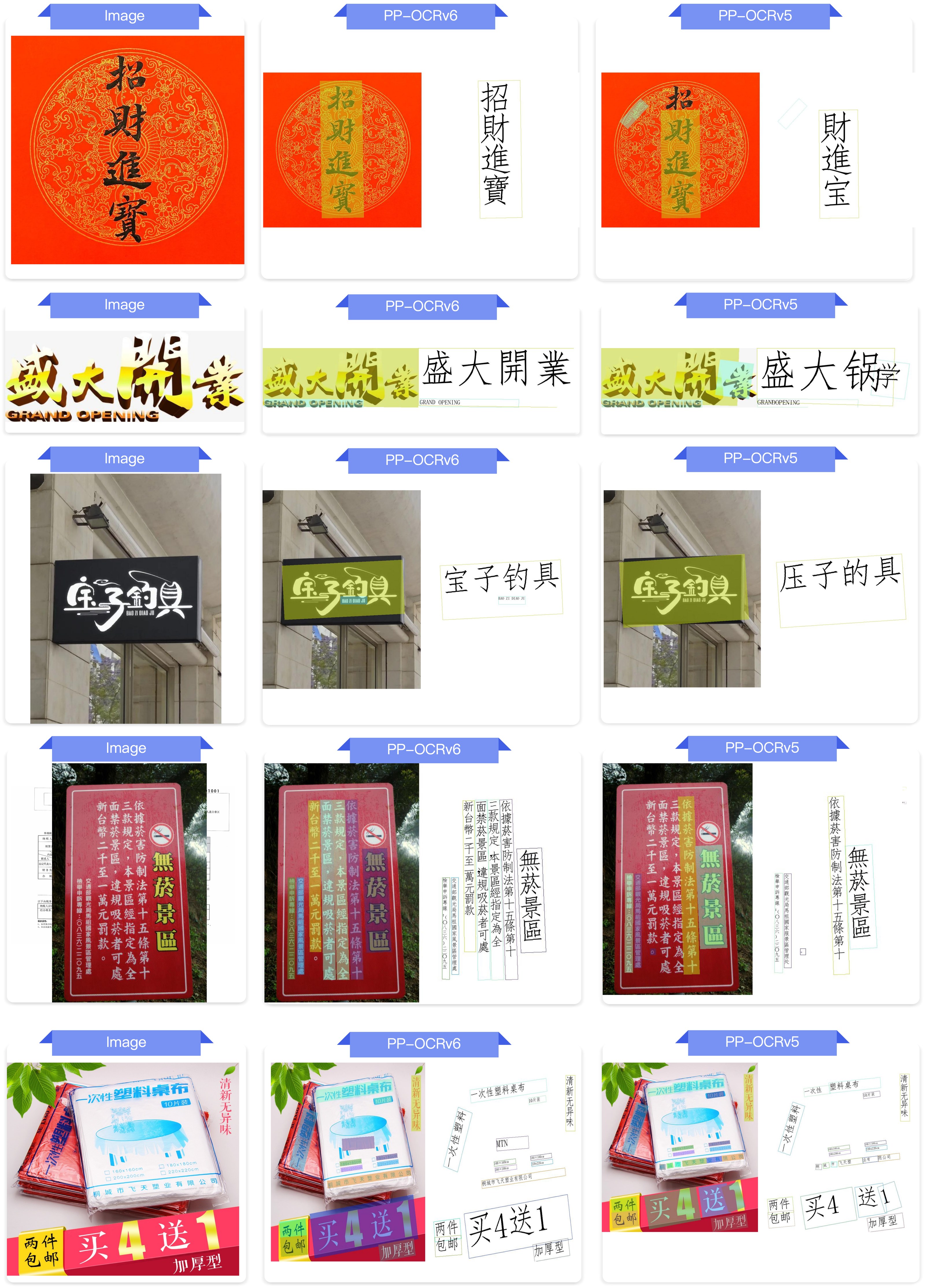}
\caption{Qualitative comparison between PP-OCRv6 and PP-OCRv5 (d).}
\label{fig:case4}
\end{figure*}

\begin{figure*}[h]
\centering
\includegraphics[width=1.0\textwidth]{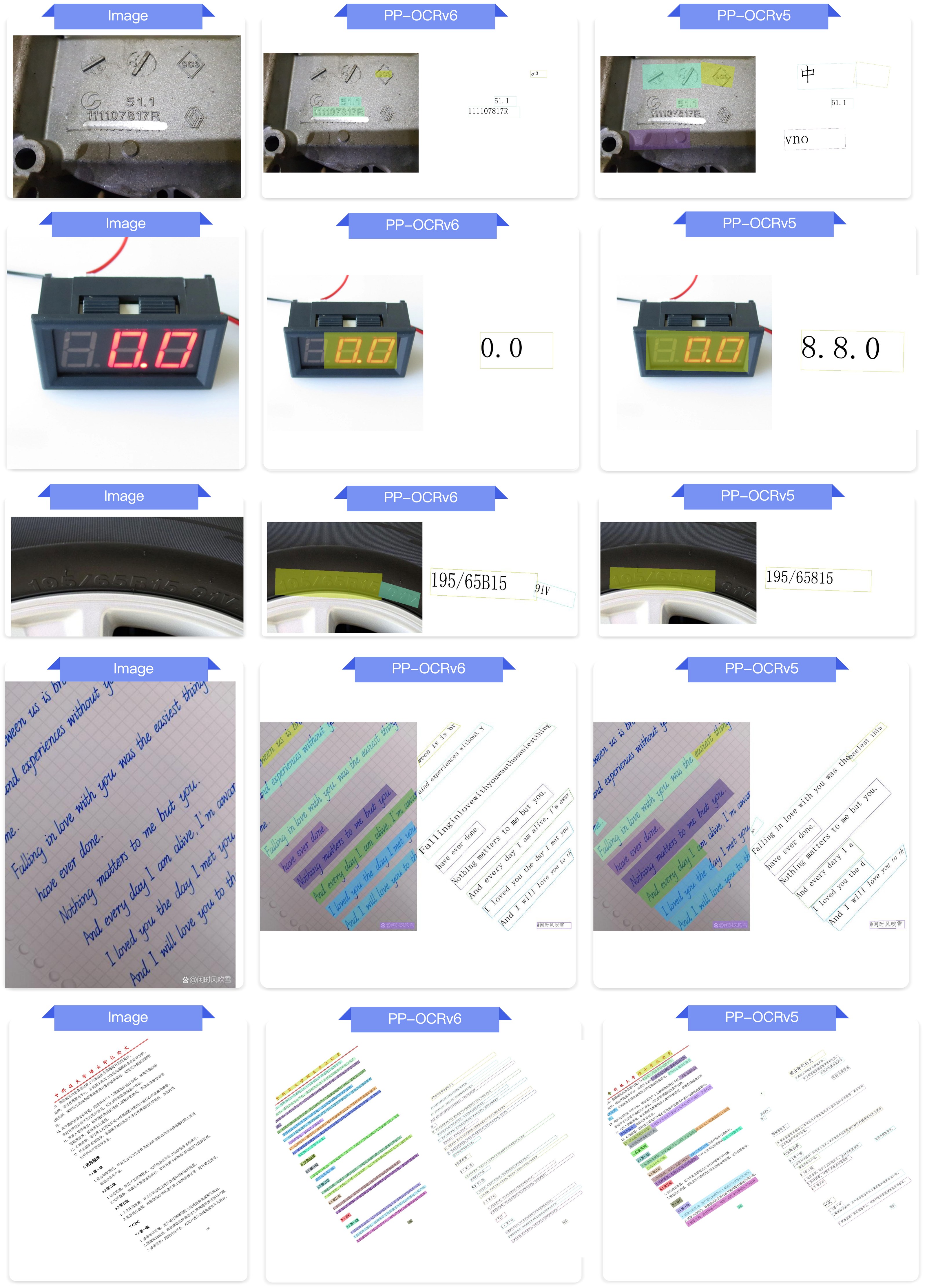}
\caption{Qualitative comparison between PP-OCRv6 and PP-OCRv5 (e).}
\label{fig:case5}
\end{figure*}

\begin{figure*}[h]
\centering
\includegraphics[width=1.0\textwidth]{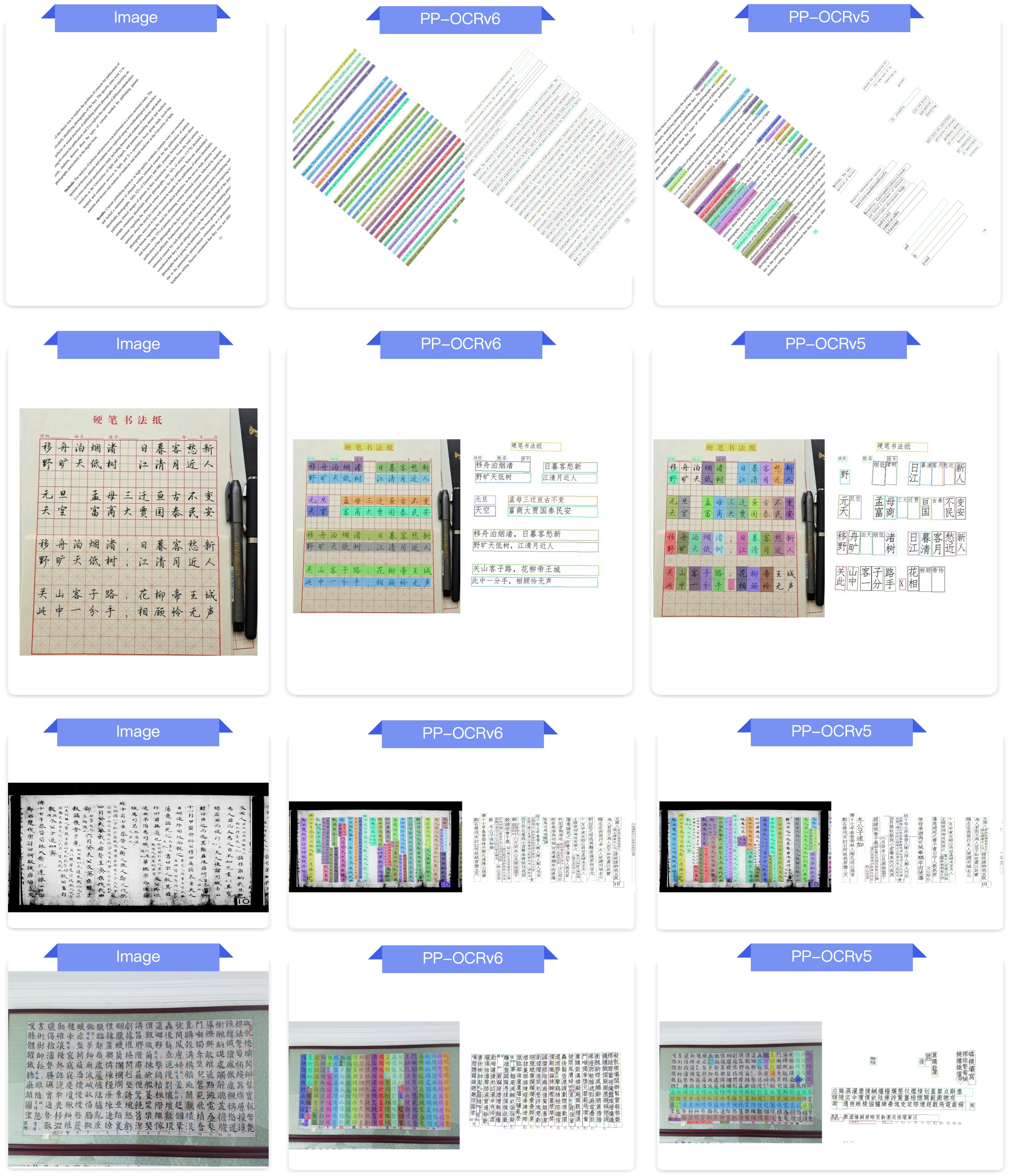}
\caption{Qualitative comparison between PP-OCRv6 and PP-OCRv5 (f).}
\label{fig:case6}
\end{figure*}

Figures~\ref{fig:case1}--\ref{fig:case6} present side-by-side end-to-end OCR results comparing PP-OCRv6\_medium and PP-OCRv5\_server across diverse real-world scenarios, covering Chinese, English, Japanese, artistic fonts, industrial characters, rotated text, pinyin, dot-matrix characters, special characters, and general scenes. Each figure shows the input image alongside the detection and recognition outputs of both systems.

% \begin{figure*}[h]
% \centering
% \includegraphics[width=1.0\textwidth]{v6_images_v2/Chinese.png}
% \caption{Qualitative comparison between PP-OCRv6 and PP-OCRv5 (g).}
% \label{fig:case7}
% \end{figure*}

% \begin{figure*}[h]
% \centering
% \includegraphics[width=1.0\textwidth]{v6_images_v2/Special and common.png}
% \caption{Qualitative comparison between PP-OCRv6 and PP-OCRv5 (h).}
% \label{fig:case8}
% \end{figure*}

% \begin{figure*}[h]
% \centering
% \includegraphics[width=1.0\textwidth]{v6_images_v2/Common.png}
% \caption{Qualitative comparison between PP-OCRv6 and PP-OCRv5 (i).}
% \label{fig:case9}
% \end{figure*}

\end{document}

%% file: commands.tex
\renewcommand{\phi}{\varphi}

\renewcommand{\epsilon}{\varepsilon}
\renewcommand{\imath}{\mathrm{i}}

\newlength{\restsubwidth}
\newlength{\restsubheight}
\newlength{\restsubmoreheight}
\newcommand{\rest}[2]{%
        \settowidth{\restsubwidth}{\ensuremath{#2}}
        \settoheight{\restsubheight}{\ensuremath{{}_{#2}}}
        \ensuremath{{#1\hskip 0.5pt}_{\vrule\kern2pt\parbox[b][%
        4pt][b]{\the\restsubwidth}{%
                        \ensuremath{{}_{#2}}}}}
        }